\documentclass{article}
\usepackage{lscape}
\usepackage{PRIMEarxiv}
\usepackage{graphicx}
\usepackage[utf8]{inputenc} 
\usepackage[T1]{fontenc}    
\usepackage{hyperref}       
\usepackage{url}            
\usepackage{booktabs}       
\usepackage{amsfonts}       
\usepackage{nicefrac}       
\usepackage{adjustbox}

\usepackage{microtype}      
\usepackage{lipsum}
\usepackage{fancyhdr}       
\usepackage{graphicx}       
\graphicspath{{media/}}     
\usepackage{amsmath}
\usepackage{graphicx}  
\usepackage{array}     

\usepackage{pifont}
\usepackage{booktabs}
\usepackage[table,xcdraw]{xcolor}
\usepackage{amssymb} 
\usepackage{multirow}
\usepackage{graphicx}
\usepackage[table,xcdraw]{xcolor}

\title{\textbf{Eyes on the Road}: State-of-the-Art Video Question Answering Models Assessment for Traffic Monitoring Tasks}

\author{
  Joseph Raj Vishal, Divesh Basina, Aarya Choudhary,  Bharatesh Chakravarthi\\
  Arizona State University \\
\texttt{jnolas77@asu.edu, dbasina@asu.edu, achoud65@asu.edu, bshettah@asu.edu} \\
}

\begin{document}
\maketitle

\begin{abstract}

Recent advances in the video question answering domain offer promising applications across various fields, particularly in traffic monitoring, where the accurate and efficient interpretation of video data is vital. Within intelligent transportation systems, the ability to answer complex, real-time queries such as \textit{“How many red cars passed in the last 10 minutes?”} or \textit{“Was there an incident between 3:00 PM and 3:05 PM?”} can significantly enhance situational awareness and decision-making. Despite notable progress in vision-language models and large language models, video question answering remains a challenging task, especially in complex, dynamic environments involving multiple objects and intricate spatiotemporal relationships.
This study presents a comprehensive evaluation of state-of-the-art video question answering models using non-benchmark video sequences comprising both synthetic and real-world traffic sequences. The evaluation framework leverages GPT-$4o$ as an objective evaluator to assess model performance based on accuracy, relevance, and consistency across three categories: basic detection, temporal reasoning, and decomposition queries. Among the models evaluated, VideoLLaMA-$2$ stood out, achieving an average accuracy of 57\%, with particular strengths in compositional reasoning and answer consistency across related queries. However, all models, including VideoLLaMA-$2$, exhibited notable limitations in multi-object tracking, temporal coherence, and complex scene interpretation, underscoring critical gaps in current video question answering architectures.
These findings highlight the potential of video question answering for traffic monitoring while emphasizing the need for further advancements. Future improvements in multi-object tracking, refined temporal reasoning, and enhanced compositional capabilities are essential for developing robust models capable of addressing the demands of real-world traffic management. By bridging these gaps, video question answering models could become indispensable tools in intelligent traffic systems, enabling faster incident detection, improved traffic flow management, and more responsive urban planning. The code and evaluation framework from this study are open-sourced for further research and exploration: 
\url{ https://github.com/joe-rabbit/VideoQA_Pilot_Study}
\end{abstract}

\section*{Keywords}
Video Question Answering, Vision-Language Models, Large Language Models, Natural Language Understanding, Spatial-Temporal Reasoning, Multi-Object Detection, Traffic Monitoring, and Traffic Scene Analysis.

\section{Introduction}
Video question answering (VideoQA) is an emerging field in artificial intelligence (AI) focusing on building models capable of answering natural language questions based on video content \cite{zhong2022videoquestionansweringdatasets}. Although intuitive for humans, VideoQA presents unique challenges for AI, requiring advanced spatial and temporal reasoning to interpret dynamic scenes. Despite advances in vision-language models (VLMs) and large language models (LLMs), achieving consistent high-performance VideoQA remains challenging, with complex interactions and sequential events \cite{xiao2024videoqaerallmsempirical}. Among various applications, traffic monitoring is a particularly relevant use case for VideoQA. AI models analyze video feeds from intersections and highways, offering real-time insights into traffic conditions, patterns, and behaviors. In this context, VideoQA supports tasks such as detection, tracking, prediction, and traffic pattern analysis to enhance safety and transportation efficiency
\cite{qasemi2023trafficdomainvideoquestionanswering}.

\begin{figure}[t]
\centering{\includegraphics[width=0.85\linewidth]{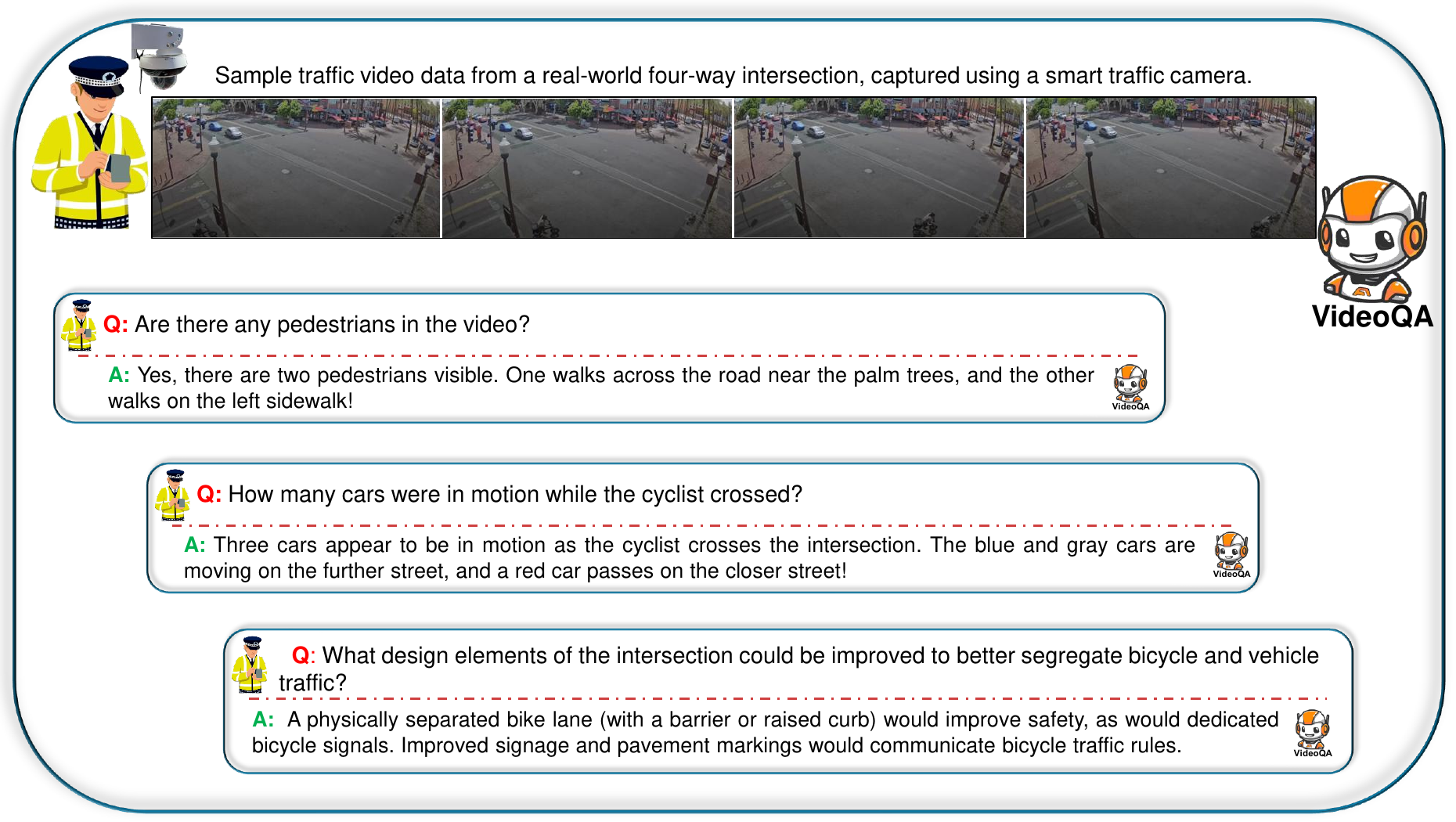}}
\caption{An example scenario of a question posed by a traffic analyst and answered by a VideoQA model is based on a traffic video sequence captured at a real-world four-way intersection.}
\label{fig01}
\end{figure}

This study evaluates the state-of-the-art (SOTA) VideoQA models specifically for traffic monitoring applications. We aim to assess models that balance high performance with computational efficiency, critical for real-time applications. Beyond basic vehicle detection, we analyze these models’ ability to interpret complex traffic scenarios, including multi-object tracking, pedestrian interactions, and compositional reasoning over time. For evaluation, we use real-world video sequences recorded from a four-way intersection (see Figure \ref{fig01}) and similar synthetic traffic data generated using CARLA simulator \cite{dosovitskiy2017carlaopenurbandriving}. Our methodology builds on the work of Marcu et al. \cite{marcu2024lingoqavisualquestionanswering}, utilizing language models as objective evaluators by leveraging natural language understanding for quality assessment. Additionally, we draw from Liao et al. \cite{liao2024alignaggregatecompositionalreasoning}, which focuses on compositional reasoning and introduces various metrics to evaluate response consistency across related queries. In this work, GPT-$4o$ \cite{gpt4o2024} is utilized as an evaluator to assess answer quality and consistency precision, forming a robust framework for evaluating the performance of VideoQA models.

The structure of this paper is organized as follows:
\begin{enumerate}
    \item \textbf{Background on VideoQA:} This section offers an overview of VideoQA, examining its spatial, and temporal dimensions, and the unique challenges it presents for AI models in processing dynamic video content.
    
    \item \textbf{Method of study:} This section outlines our criteria for selecting videos, focusing on the use of non-benchmark datasets that combine synthetic and real-world footage. We also describe the question generation and filtering process, ensuring questions are both relevant and challenging for model evaluation.
    
    \item \textbf{Rationale of SOTA VideoQA models:} This section presents  SOTA models chosen for the study, their architecture, and selection criteria supported by a comprehensive comparison, highlighting each model's capabilities.
    
    \item \textbf{Evaluation methodology:} This section details the metrics and methods used to evaluate model performance, including accuracy scores based on alignment with ground-truth answers and compositional consistency metrics like accuracy and consistency precision (cP).
    
    \item \textbf{Integration of prior methodologies:} This section explains how the evaluation frameworks from Marcu et al. \cite{marcu2024lingoqavisualquestionanswering}  and Liao et al. \cite{liao2024alignaggregatecompositionalreasoning}
    were combined to create a unified approach. This integrated framework leverages objective scoring alongside compositional consistency metrics, offering a more comprehensive evaluation.
\end{enumerate}

The rest of the paper provides a systematic exploration of VideoQA methodologies, from foundational concepts to detailed evaluations. By combining insights from prior work, dataset selection, and an evaluation framework, this study aims to offer a comprehensive analysis of SOTA  VideoQA models. The findings and comparisons presented in the subsequent sections will contribute to advancing the field by addressing existing challenges and highlighting potential directions for future research.

\section{VideoQA Foundations for Traffic Monitoring Applications} \label{foundation-videoQA}

This section introduces the fundamentals of VideoQA, outlining its unique challenges, particularly in traffic monitoring applications. Additionally, it describes the data sources used in this study involving real-world and simulated traffic video sequences to evaluate VideoQA models.

\subsection{Core Challenges in VideoQA}

VidedvzdoQA is an interdisciplinary field that combines computer vision, natural language processing, and multimodal learning to create systems capable of answering natural language questions based on video content \cite{zhong2022videoquestionansweringdatasets}. Unlike traditional image-based question answering, VideoQA must interpret both spatial and temporal information, requiring models to grasp not only individual objects and actions but also their evolving dynamics across video sequences \cite{10582849}. The evolution of VideoQA \cite{gan2016recognizing,zhu2019video,xu2017video} is rooted in advancements in deep learning, particularly the application of convolutional neural networks (CNNs) for visual feature extraction and transformers for processing sequential data. Early VideoQA models \cite{NEURIPS2019_c74d97b0, 8839734} primarily relied on pre-trained vision and language models, which limited their ability to perform temporal reasoning and constrained them to simpler queries. Models like VisualBERT \cite{li2019visualbertsimpleperformantbaseline} integrated vision and language for static images, while recent vision-language pre-trained architectures have broadened VideoQA capabilities, allowing models to learn complex correspondences between video sequences and natural language, thereby supporting both spatial and temporal reasoning.
Several datasets are commonly used to test VideoQA models across diverse tasks, including activity recognition, temporal reasoning, and scene comprehension. 
Typically, VideoQA models are trained and tested on common benchmark datasets that include: 

\begin{itemize} 

\item \textbf{ActivityNet-QA} \cite{7298698}: A large-scale dataset designed to test VideoQA models on human activity recognition and complex action-oriented questions. 
\item \textbf{TGIF-QA} \cite{li2016tgifnewdatasetbenchmark}: A dataset focused on animated GIFs, presenting questions that require models to perform temporal reasoning tasks, such as counting repeated actions. 
\item \textbf{MSRVTT-QA} \cite{xu2017video}: This dataset features diverse, real-world video content, and challenging models with complex, scene-based queries.
\item \textbf{NExT-QA} \cite{xiao2021nextqanextphasequestionansweringexplaining}: A VideoQA dataset that includes questions requiring event sequence comprehension, reasoning about interactions, and causal relationships.
\item \textbf{1H-VideoQA Dataset} \cite{geminiteam2024geminifamilyhighlycapable}: A benchmark dataset designed for long-form video comprehension, featuring videos up to 10.5 hours, challenging models to retrieve accurate information across extensive temporal spans.
\item \textbf{LLaVA-Video-178K} \cite{zhang2024videoinstructiontuningsynthetic}: A dataset for advanced video comprehension, containing clips from large video collections, focusing on detailed event understanding. It was designed using GPT-$4o$.
\item \textbf{Ego4D} \cite{grauman2022ego4dworld3000hours}: A large-scale dataset of egocentric (first-person) videos that captures a wide range of daily activities, used to train models for tasks like action recognition and event detection from a first-person perspective.
\item \textbf{EgoSchema} \cite{egoschema}: A dataset focused on short egocentric videos, designed to test models on precise scene understanding and VideoQA tasks in ego-centric contexts.
\end{itemize}

Despite being trained on extensive datasets, VideoQA models still face significant challenges, such as:

\begin{itemize} 
\item \textbf{Temporal and spatial reasoning}: Models often struggle with spatial  (understanding relationships within frames) and temporal reasoning (tracking changes across sequences) \cite{gpt4o2024,rawal2024cinepile,xiao2024videoqaerallmsempirical}.

\item \textbf{Complex interaction parsing}: Real-world scenarios often contain multiple interactions occurring simultaneously, requiring models to differentiate and accurately interpret the dynamics \cite{Rahman_2024_WACV, NIPS2014_d516b136, liao2024alignaggregatecompositionalreasoning,xiao2024videoqaerallmsempirical}. 

\item \textbf{Compositional reasoning}: Many VideoQA tasks require compositional reasoning, where models must integrate information from different video segments to answer multi-step questions that involve both foreground and background elements \cite{rawal2024cinepile, NIPS2014_d516b136,geminiteam2024geminifamilyhighlycapable,xiao2024videoqaerallmsempirical}. 

\item \textbf{Efficiency constraints}: Real-time applications, such as traffic monitoring, demand models that combine accuracy with computational efficiency, presenting a significant design challenge for VideoQA systems \cite{NIPS2014_d516b136,awad2023trecvid, Xu_2021_CVPR, xiao2024videoqaerallmsempirical}. 
\end{itemize}

Although advances in VLMs and LLMs have expanded the capabilities of VideoQA, The performance is still hampered by the variability of real-world video data, which includes diverse resolutions, lighting conditions, and object interactions. Research efforts continue to focus on developing models that generalize well across different video contexts, to create robust, adaptable VideoQA systems.
This study specifically examines the performance of SOTA VideoQA models to handle various downstream tasks specific to traffic monitoring.

\subsection{Traffic Video Data: Sources and Description}

To thoroughly evaluate VideoQA models' performance, we use traffic video data from two key sources: real-world sequences collected from a traffic intersection in Tempe, AZ, and similar traffic data generated using the CARLA simulator.
\begin{figure}[t]
\centering{\includegraphics[width=1\linewidth]{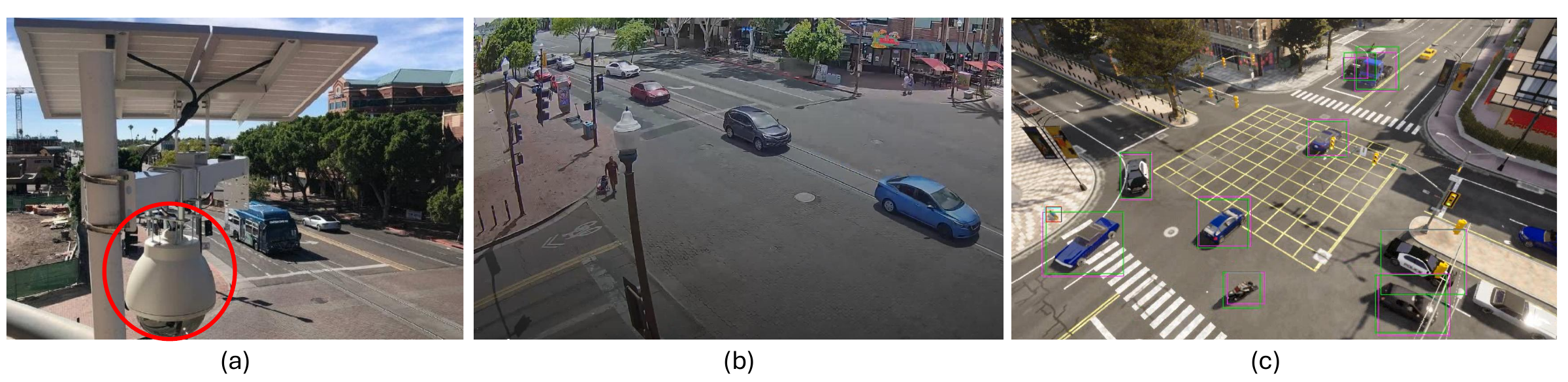}}
\caption{Traffic Video Data Sources - (a) Real-world traffic video captured at a four-way intersection in Tempe, AZ. (b) Sample video frame from the Tempe intersection, showing real-world traffic interactions. (c) Simulated video frame from CARLA, depicting a four-way intersection with varied vehicle and pedestrian activity in a controlled environment.}
\label{fig02}
\end{figure}
The real-world traffic video data was collected in collaboration with Argos Vision \cite{argosvision}, an AI-based smart camera solution provider, using a monocular RGB camera mounted at a four-way intersection in Tempe, AZ (See Figure \ref{fig02} (a)). This camera setup captures video sequences at various times of day and in varying durations, ensuring a representative sample of different traffic patterns and interactions. To create a dataset relevant to traffic monitoring tasks, we carefully selected sequences containing incidents and events of interest to traffic engineers. These include situations that highlight complex traffic behavior, such as red-light violations, vehicle counting (See Figure \ref{fig02} (b)), and other events that might require intervention or attention in real-world traffic management.

To expand the scope of evaluation and test the models under controlled yet realistic scenarios, we also used the CARLA simulator \cite{dosovitskiy2017carlaopenurbandriving} to create simulated traffic data. The CARLA simulation replicates a similar four-way intersection setting, incorporating vehicles, pedestrians, and micromobility elements under diverse conditions (See Figure \ref{fig02} (c)). This simulated data includes variations in lighting, weather, and vehicle behaviors, providing a comprehensive testing ground for VideoQA models. These environmental changes help to challenge the models, ensuring they are capable of adapting to different visual and contextual cues.
By leveraging both real-world and simulated data, this study emulates the tasks of a traffic engineer utilizing  VideoQA models’ capabilities to provide accurate, real-time insights into complex traffic scenes.

\section{Method of Study} 
\label{method-of-study}

This section outlines the methods used in the study, detailing the selection of video sequences, the process for generating and curating questions, and the filtering criteria applied to ensure alignment with the study's objectives. Video selection prioritized non-benchmark datasets featuring short, multi-object videos of real-world scenarios to evaluate model generalization. Questions were organized into three levels of difficulty - easy, medium, and complex to assess model capabilities in object identification, temporal awareness, and handling of negation. Each question was carefully designed and filtered to ensure content relevance and alignment with the study’s goals, effectively testing the VideoQA models’ observational and reasoning abilities.

\subsection{Video Data Selection} 

The primary criterion for video selection was to use non-benchmark datasets, as most SOTA models have already been extensively trained on popular datasets such as ActivityNet \cite{7298698}, TGIF \cite{li2016tgifnewdatasetbenchmark}, and MSRVTT \cite{xiao2024can}. The goal was to evaluate these models' generalization capabilities by testing them on unseen data. For example, in one of the evaluations, a model was given a traffic surveillance video and was tasked with performing multi-object detection and temporal information analysis. To maintain fairness, particularly for models not designed for long-form VideoQA (e.g., videos exceeding one hour), the selected videos were kept between $10$ to $45$s in length. This time frame ensures that each model can process the videos without scalability issues, allowing for a consistent and comparable evaluation. As mentioned in Section \ref{foundation-videoQA}, traffic videos, both synthetic and real-world (see Figure \ref{fig03}), were chosen as the primary data sources due to their relevance and varied quality. Videos were carefully selected to maintain human-viewable resolutions and consist of scenarios involving multiple objects and complex interactions, which require strong spatiotemporal reasoning capabilities from the models.
We summarize the real-world and synthetic video sequences used in this study to challenge VideoQA models in diverse, realistic traffic scenarios as follows:

\begin{figure}[t]
\centering{\includegraphics[width=0.95\linewidth]{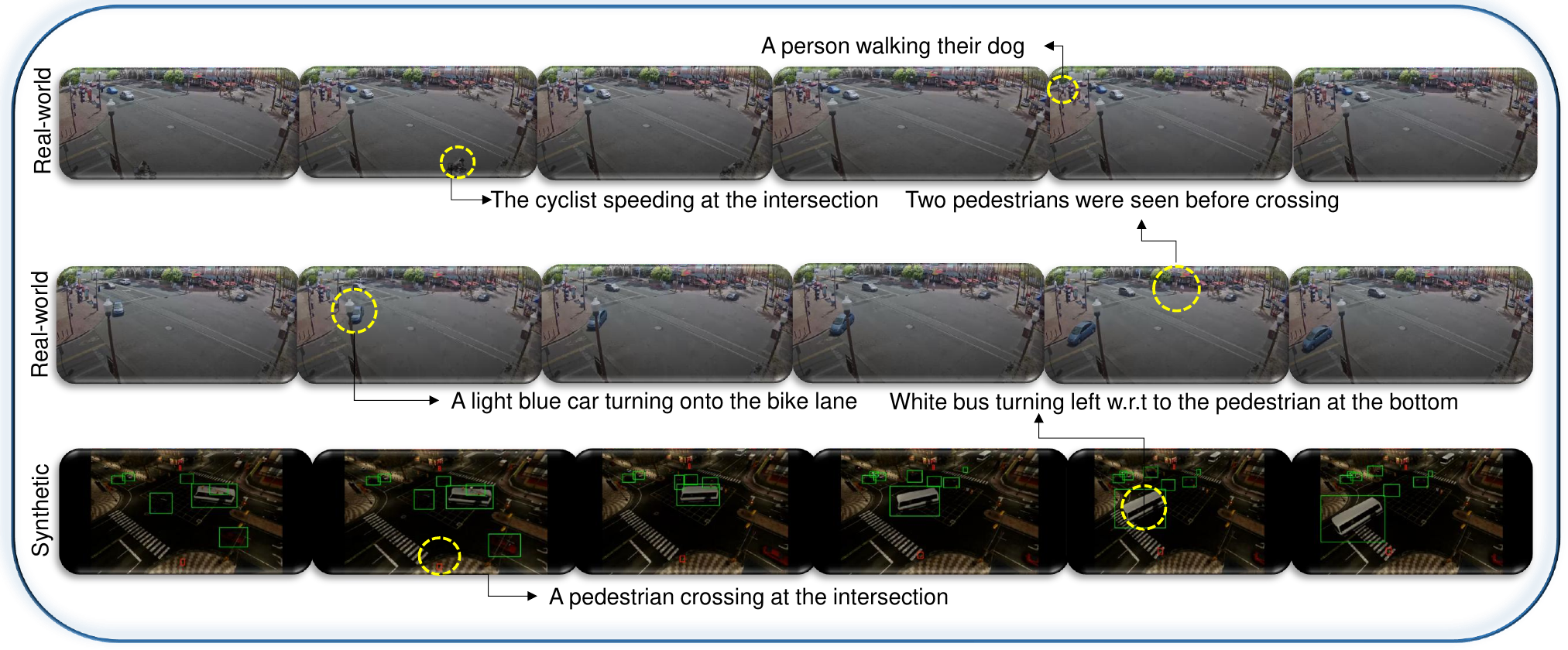}}
\caption{Sample Scenarios from traffic Videos - Sequence 1 (top) represents a real-world scenario captured at a four-way intersection on S. Mill Avenue in Tempe, AZ, featuring a cyclist. Sequence 2 (middle) depicts cars turning onto a bike lane, captured in another real-world setting. Sequence 3 (bottom) is synthesized using CARLA, illustrating a simulated four-way intersection with diverse vehicle and pedestrian activity.}
\label{fig03}
\end{figure}

\begin{enumerate}  \label{real-world-sequences}
\item \textbf{Real-world sequences}: The video sequences depict a busy intersection in Tempe, AZ, and feature various complex, real-world events. The following two real-world sequences were selected for evaluating the VideoQA models: 
\begin{enumerate}
\item \textbf{Sequence 1}: A sequence of $15$s features a cyclist speeding on the wrong side of the road, making a rapid turn without stopping, with multiple pedestrians and vehicles moving through the crosswalk (Figure \ref{fig03} top row). This scene challenges the models with simultaneous object tracking in a high-density environment.
\label{cyclist}
\item \textbf{Sequence 2}: A sequence of $45$s features four cars turning into a cycle lane. The vehicles include a sky-blue hatchback, a dark-blue, maroon sedan, and a white SUV, with four pedestrians visible with faces obscured for privacy (Figure \ref{fig03} middle row). This scene was selected to assess the models' ability to perform multi-object detection and analyze intricate interactions.
\end{enumerate}
\label{wrong_turn}    
\item \textbf{Synthetic sequence} (Sequence 3): A sequence of $10$s depicting a nighttime intersection scene where fog gradually sets in, with a pedestrian crossing from the bottom left to the bottom right of the frame as a yellow taxi stops to allow passage (Figure \ref{fig03} bottom row). This scenario was included to evaluate the models' ability to handle synthetic environments and detect specific actions.
\label{synthetic}   
\end{enumerate}

These concise, multi-object videos provide an ideal setting for evaluating the spatio-temporal reasoning abilities of current SOTA VideoQA models.

\subsection{Structured Question Generation} \label{structured-questions}

\begin{table}[]
\caption{Sample questions categorized by difficulty level for SOTA VideoQA models performance analysis.}
\label{table01}
\resizebox{\columnwidth}{!}{%
\begin{tabular}{|
>{\columncolor[HTML]{EFEFEF}}l 
>{\columncolor[HTML]{FFFFFF}}l }
\toprule \hline
\multicolumn{2}{c}{\cellcolor[HTML]{EFEFEF}\textbf{Sample Questions used  to evaluate the SOTA VideoQA models}} \\ \hline 
\multicolumn{1}{c|}{\cellcolor[HTML]{EFEFEF}\textbf{Category}} &
  \multicolumn{1}{c}{\cellcolor[HTML]{EFEFEF}\textbf{Questions}} \\ \hline \hline
\multicolumn{1}{c|}{\cellcolor[HTML]{EFEFEF}} & Is the pedestrian using a crosswalk?                             \\ \cline{2-2} 
\multicolumn{1}{c|}{\cellcolor[HTML]{EFEFEF}} & How many yellow cars are there?                                  \\ \cline{2-2} 
\multicolumn{1}{c|}{\cellcolor[HTML]{EFEFEF}} & How many traffic lights are visible?                             \\ \cline{2-2} 
\multicolumn{1}{c|}{\multirow{-4}{*}{\cellcolor[HTML]{EFEFEF}\textbf{Easy}}} &
  \textbf{Negated Question}: Is the pedestrian not using a crosswalk? \\ \hline \hline
\multicolumn{1}{l|}{\cellcolor[HTML]{EFEFEF}} & How many pedestrians cross the road in total?                    \\ \cline{2-2} 
\multicolumn{1}{l|}{\cellcolor[HTML]{EFEFEF}} & Was the pedestrian crossing from left to right or right to left? \\ \cline{2-2} 
\multicolumn{1}{l|}{\cellcolor[HTML]{EFEFEF}} & Which car arrived first at the intersection?                     \\ \cline{2-2} 
\multicolumn{1}{l|}{\multirow{-4}{*}{\cellcolor[HTML]{EFEFEF}\textbf{Medium}}} &
   \textbf{Negated Question}: How many pedestrians did not cross the road? \\ \hline \hline
\multicolumn{1}{l|}{\cellcolor[HTML]{EFEFEF}} & How many vehicles stopped after turning?                         \\ \cline{2-2} 
\multicolumn{1}{l|}{\cellcolor[HTML]{EFEFEF}} & Was traffic light timing optimal?                                \\ \cline{2-2} 
\multicolumn{1}{l|}{\cellcolor[HTML]{EFEFEF}} & Between 0:08 and 0:10, how many cars are moving?                 \\ \cline{2-2} 
\multicolumn{1}{l|}{\multirow{-4}{*}{\cellcolor[HTML]{EFEFEF}\textbf{Complex}}} &
  \textbf{Negated Question}: Vehicles not in the middle of an intersection after pedestrian crossed? \\ \hline \bottomrule
\end{tabular}%
}
\end{table}

Inspired by compositional reasoning techniques outlined in \cite{liao2024alignaggregatecompositionalreasoning}, the question-generation process in this study was designed to maximize the extraction of spatial and temporal information from selected videos. Building upon the concept of question decomposition trees, which break down complex queries into simpler components, this approach expanded the idea by generating multiple questions that vary in complexity and reasoning requirements. Each question was crafted to target distinct reasoning abilities, including identification, temporal understanding, and negation handling, aligning with the principles of compositional reasoning. Questions were categorized into three levels - easy, moderately challenging, and complex - to emulate the decomposition process across a broader spectrum. Instead of a single decomposition tree breaking down a single complex query, each question acted as a branch exploring different aspects of the video content, facilitating a thorough evaluation of the model's spatial, temporal, and reasoning capabilities. This process generated questions across three difficulty levels, designed to prompt a range of reasoning abilities, from straightforward identification to in-depth scenario analysis.
We describe the question categories below, and Table \ref{table01} presents a sample of the questions used in the study, organized by difficulty level. The full set of questions is available in the supplementary material.

\begin{itemize}
\item \textbf{Easy questions}: These questions are basic yes/no or counting-based questions that require minimal inference (e.g., ``Is there a person walking a dog?” based on Real-World Video Sequence $1$). Such questions help gauge the model's ability to identify straightforward objects and actions.
    
\item \textbf{Moderately hard questions}: Questions in this category require models to provide an explanation of the setting and events occurring within specific timeframes. This type of question tests the model's temporal reasoning skills and ability to contextualize actions within the video sequence.
    
\item \textbf{Complex questions}: Complex questions introduce elements of negation or ask about events that do not occur in the video (e.g., ``Is there a truck present?”). This level of questioning evaluates the model’s tendency to hallucinate and tests its capacity to differentiate between present and missing elements.
\end{itemize}

The question generation process employed GPT-$4o$ \cite{gpt4o2024}, which was prompted with a specialized template (included in the appendix). This prompt included human-generated examples to set a quality standard and guide GPT-$4o$ \cite{gpt4o2024} in generating questions that aligned with human reasoning patterns. A total of $18$ questions for each video was curated with $5$ questions $+$ $1$ negated questions under each category.

\subsection{ Questions Selection Criteria}

We initially had a pool of GPT-$4o$-generated questions that aligned with human reasoning patterns. However, to ensure consistency, relevance, and alignment with the study's objectives, the questions were carefully filtered and selected based on a set of defined criteria \cite{liao2024alignaggregatecompositionalreasoning}. These criteria were designed to extract meaningful insights from the VideoQA models while testing various aspects of their reasoning and observational capabilities. The selected criteria are as follows:

\begin{enumerate}
\item \textbf{Appropriate question types}:  
The question types were deliberately chosen to include only those that could be answered using observable data, such as yes/no, how, where, and what questions. ``Why” questions were excluded, as they typically require inferential reasoning or external knowledge beyond the scope of direct video analysis. This ensured that all questions could be answered based solely on the visual and temporal information available in the video. For instance, questions like ``Where is the pedestrian?” or ``What color is the car?” were included, while ``Why is the person crossing the street?” were excluded.
    
\item \textbf{Relevance to video content}:  
Each question was crafted to directly relate to the events, objects, and interactions shown in the video. This approach ensured that the questions were relevant and contextually anchored in the visual content, maintaining focus on the specific elements depicted. Irrelevant questions were avoided to ensure clarity and prevent confusion for the model. For example, in a traffic video, questions would address vehicles, pedestrians, or actions within the scene, rather than unrelated topics.
    
\item \textbf{Quantitative questions}:  
These were included to assess the model’s ability to count objects or actions within the video. These questions required the model to track and enumerate visible elements, testing its spatial awareness and attention to detail. For example, questions like ``How many cars are visible?” or ``How many pedestrians cross the intersection?” evaluate the model's capacity to monitor multiple entities and provide accurate counts, a crucial skill for interpreting complex scenes.
    
\item \textbf{Implied temporal data}:  
Instead of explicitly requesting timestamps, questions were designed to imply temporal understanding. This approach evaluated the model's ability to interpret events or actions in a sequential context without referring to specific timeframes. For example, questions like ``Does the cyclist pass the pedestrian before the car arrives?” test the model's ability to infer temporal relationships, focusing on relative timing rather than precise timestamps.

\item \textbf{Single-part focus}:  
To ensure clarity and simplicity, questions were designed to focus on a single aspect of the video. This approach avoided multi-part questions that could create ambiguity or complicate interpretation. By isolating one element at a time, the questions enabled more accurate assessments of the model’s ability to focus on specific observations. For example, instead of asking, ``What vehicles and pedestrians are present, and what are they doing?” the questions were separated into ``What vehicles are present?” or ``What are the pedestrians doing?”

\end{enumerate}

To ensure the effectiveness of these criteria, each question was carefully reviewed for clarity, relevance, and alignment with the study’s objectives. By adhering to these guidelines, we aimed to create a well-defined set of questions that would provide meaningful insights into the capabilities of the VideoQA models, testing their ability to understand and reason about complex, real-world traffic scenarios. The selected questions were then used to evaluate the models across various visual perception, temporal reasoning, and contextual understanding aspects.

\section{SOTA VideoQA Models and the Selection Rationale}
\label{sect_04}
This section presents the SOTA VideoQA models evaluated in this study. The selected models include open-source  (LLAVA \cite{li2024llava,li2024llavanext}, InternVL \cite{chen2024far}, and VideoLLaMA \cite{cheng2024videollama2advancingspatialtemporal}) and proprietary models (GPT-$4o$ \cite{gpt4o2024} and Gemini-$1.5$ \cite{geminiteam2024geminifamilyhighlycapable}).
These models were chosen for their advanced multimodal capabilities, particularly their ability to effectively integrate video, audio, and text inputs for effective VideoQA. The selection was guided by the goal of capturing a diverse range of features and performance characteristics, leveraging each model's unique strengths to address the specific demands of VideoQA tasks. The following subsections highlight the distinctive attributes and the rationale for their inclusion in this study.

\subsection{Internal Vision-Language (Intern-VL) }

\begin{figure}[t]
\centering{\includegraphics[width=1\linewidth]{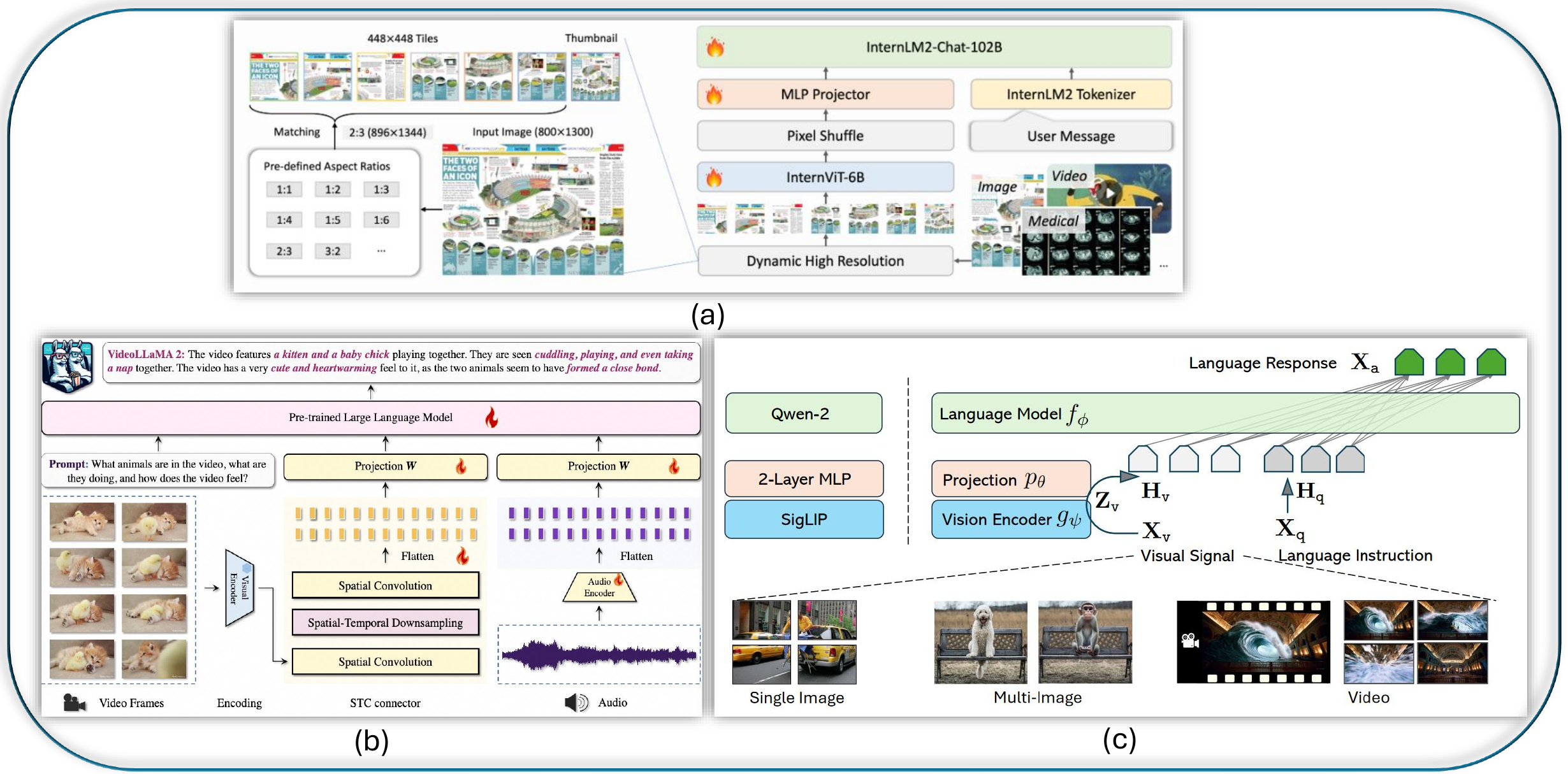}}
\caption{Shows the architectures of three open-source models: (a) InternVL \cite{chen2023internvl,chen2024far}, (b) VideoLLaMA-2 \cite{damonlpsg2023videollama}, and (c) the general architecture of the LLAVA-One family \cite{li2024llava}.}
\label{fig04}
\end{figure}

The Intern-VL \cite{chen2024far} family of models is an open-source suite designed to integrate both visual and textual information for advanced multimodal tasks. Built on a robust vision transformer–multi-layer perceptron–large language model (ViT-MLP-LLM) architecture \cite{dong2024internlm, liu2023visualinstructiontuning, lu2021interpretablegeometry}, these models excel in processing high-resolution visual content alongside natural language inputs, making them highly versatile for vision-language tasks. The Intern-VL architecture (See Figure \ref{fig04} (a)) comprises:

\begin{itemize}
\item \textbf{Vision encoder} - The internViT-6B module \cite{chen2023internvl} is a transformer-based image encoder with a high-resolution capacity, able to process images up to 4K px. It divides images into ${448\times448}$ px segments to maintain detailed spatial context, supporting tasks involving complex image structures, such as document interpretation and detailed scene analysis.
\item \textbf{Language model} - The internLM2-20B model \cite{cai2024internlm2} processes natural language data and generates responses based on textual and multimodal inputs. Optimized for tasks like dialogue and question answering, it enables accurate interpretation of multimodal inputs within complex linguistic contexts.
\item \textbf{MLP transformer} - It integrates the output from the vision encoder to the language model, transforming visual embedding into a format compatible with the language model’s input. This transformation involves several key steps:
\begin{itemize}
\item \textbf{Dimensional adjustment}: Adapts the dimensionality of visual features to align with the language model's embedding requirements.
\item \textbf{Non-Linear transformation}: Uses non-linear functions, like rectified linear unit (ReLU) activation, to capture complex relationships between visual and textual data.
\item \textbf{Feature integration}: Ensures that visual features are effectively incorporated into the language model’s processing pipeline, allowing it to generate responses that accurately reflect visual context.
\end{itemize}
\end{itemize}

Furthermore, The Intern-VL models were trained on a comprehensive, high-quality bilingual dataset across several task categories, including general question answering, scientific document analysis, chart interpretation, and optical character recognition (OCR). The training datasets span both English and Chinese, supporting robust OCR and document understanding in multiple languages. Evaluation was performed on specific benchmarks, including DocVQA for document comprehension \cite{mathew2021docvqa}, ChartQA for chart understanding \cite{masry2022chartqa}, InfographicVQA for infographic interpretation \cite{mathew2022infographic}, TextVQA for scene text analysis \cite{singh2019towards}, OCRBench \cite{liu2023ocrbench}, and the MMT-Benchmark dataset for visual question answering \cite{mmtbench}.
However, Intern-VL $2$ has high computational requirements, especially for high-resolution image processing, which may limit its use in environments with constrained resources. Furthermore, its alignment process, which is tailored for structured data, may not perform as well with unstructured visual inputs, reducing its effectiveness in tasks that deviate from well-defined document or scene layouts.

In this study, particularly the InternVL2-8B variant \cite{chen2023internvl,chen2024far}, was chosen due to its ability to handle high-resolution structured data, making it a strong fit for VideoQA tasks involving documents, charts, or complex visual-text integration. This model, being comparatively light in computational requirements, was selected for its efficiency and adaptability in handling multimodal data.

\subsection{VideoLLaMA-$2$}

VideoLLaMA-$2$ \cite{cheng2024videollama2advancingspatialtemporal} is an open-source model designed to handle video and audio data with advanced spatiotemporal integration capabilities. With a dual-branch architecture featuring separate pathways for vision and audio processing, it is particularly effective for video-language tasks such as VideoQA.
Several unique features set VideoLLaMA-$2$ apart. Central to its design is the spatial-temporal convolution (STC) connector, which uses $3$D convolutions and RegStage blocks to efficiently process spatial and temporal data, reducing token overload while preserving critical visual details across frames. To ensure precise alignment of visual and auditory inputs, the model incorporates high-quality audio-visual synchronization through the BEATs encoder and a cohesive MLP projection layer. Additionally, its instruction-following capabilities, enabled by a Mistral-Instruct-tuned LLM, allow it to handle complex tasks requiring detailed and contextually relevant responses. The architecture (See Figure \ref{fig04} (b)) of VideoLLaMA-$2$ comprises three key components:

\begin{itemize}
\item \textbf{Vision-Language branch}: The visual processing pathway utilizes the CLIP model (ViT-L/$14$) \cite{radford2021learning} as its vision encoder, which processes individual video frames and extracts frame-level visual features. VideoLLaMA-$2$ introduces the STC connector, which replaces the Q-Former \cite{damonlpsg2023videollama} used in its predecessor. It further integrates spatiotemporal information using RegStage blocks for spatial refinement and $3$D convolutions for temporal aggregation, effectively reducing the number of spatial-temporal tokens and improving processing efficiency while retaining key details.

\item \textbf{Audio-Language branch}: Audio processing is handled by the BEATs encoder \cite{chen2023beats}, adept at converting audio inputs into log-mel spectrograms to capture temporal audio features. These features are then projected through an MLP layer for alignment with the visual features. This branch allows VideoLLaMA-$2$ to incorporate nuanced auditory information, enriching tasks requiring visual and audio context.

\item \textbf{LLM backbone}: The LLM component in VideoLLaMA-$2$, fine-tuned using Mistral-Instruct \cite{jiang2023mistral7b}, integrates inputs from both the visual and audio branches. This setup enables the model to analyze and respond to complex scenarios in videos where visual and auditory cues interplay, enhancing performance on tasks requiring comprehensive scene understanding.
\end{itemize}

VideoLLaMA-2 is trained on diverse large-scale datasets to ensure robust multimodal capabilities. It leverages video-language datasets such as Panda-$70M$ \cite{chen2024panda70m}, VIDAL-$10M$ \cite{zhu2023languagebind}, WebVid-$10M$ \cite{bain2021frozen}, and InternVid-$10M$ \cite{wang2023internvid}, providing extensive video-text pairs. Additionally, image-text datasets like CC-$3M$ \cite{changpinyo2021conceptual} and DCI \cite{urbanek2023picture} enhance their ability to interpret static visual data, further improving their multimodal performance.
However, the model has some limitations. Its reliance on pre-trained encoders may limit adaptability across diverse datasets, and the STC connector’s computational demands could affect performance when processing very long video sequences or subtle audio cues.
Despite these challenges, VideoLLaMA-$2$ introduces significant advancements in multimodal video processing. Its dual-branch architecture ensures seamless integration of visual and auditory information, while the STC connector optimizes spatiotemporal token management. These innovations make it a powerful tool for VideoQA and other video-language tasks, pushing the boundaries of multimodal research.

\subsection{Large Language Vision Assistant (LLaVA) }

The LLaVA \cite{li2024llava}, which has evolved into LLaVA-Next-Video \cite{zhang2024videoinstructiontuningsynthetic}, is an open-source model specifically designed for advanced video comprehension tasks.  By leveraging video-specific datasets such as LLaVA-Video-$178K$ \cite{zhang2024videoinstructiontuningsynthetic}, derived from HD-VILA-$100M$ \cite{xue2022hdvila} and Ego$4D$ \cite{grauman2022ego4dworld3000hours}, the model achieves robust multimodal understanding across video, audio, and text inputs. The key features of LLaVA-Next-Video address challenges in VideoQA and other video-language tasks:

\begin{itemize}
\item \textbf{Needle-in-a-Haystack capability}: The model is designed to retrieve specific details from within large-scale contexts, a feature critical for long-form content processing in VideoQA tasks.
\item \textbf{Multimodal processing}: It supports multimodal input processing (text, audio, and video), allowing it to interpret and align information across these modalities effectively.
\item \textbf{Hierarchical annotation framework}: The video segments are annotated with hierarchical descriptions ($10s$, $30s$, end of video), which supports a layered understanding of events over time.
\end{itemize}

The architecture of LLaVA-Next-Video (See Figure \ref{fig04} (c)) comprises several components as summarized below, each of which is designed to support the model’s ability to process and understand complex video data, as inspired by LLaVA-One \cite{li2024llava}:
\begin{itemize}
\item \textbf{Vision encoder}: A pre-trained CLIP model (ViT-L/$14$) \cite{radford2021learning} is used to extract visual features from individual video frames. This encoder, consistent with LLaVA-One's focus on efficient visual feature extraction, enables precise identification of objects and interactions within video frames.
\item \textbf{Language model}: The Vicuna language model \cite{vicuna2023} is utilized for language processing, allowing the model to interpret captions, answer questions, and generate explanations based on video content.
\item \textbf{Projection layer}: This layer aligns high-dimensional visual features with the input format of the language model, allowing seamless integration of visual and textual information for unified scene comprehension.
\item \textbf{Interleaved image-text approach}: Inspired by the image-text handling strategy of LLaVA-One, LLaVA-Next-Video \cite{zhang2024videoinstructiontuningsynthetic} interleaves visual tokens (video frames) with text tokens (such as captions or questions), enabling simultaneous capture of spatial and temporal information.
\end{itemize}

LLaVA-Next-Video demonstrates notable strengths, particularly in processing spatiotemporal data for tasks requiring sequence comprehension, scene analysis, and interaction recognition. These capabilities have been validated on VideoQA benchmarks such as NExT-QA \cite{xiao2021nextqanextphasequestionansweringexplaining} and ActivityNet-QA \cite{7298698}, where the model’s memory architecture and multimodal integration contribute to its effectiveness.
Despite its advantages, 
LLaVA-Next-Video model faces some challenges. 
High computational demands can arise from processing extensive video and audio inputs, and its reliance on dataset diversity impacts performance when dealing with rare or complex interactions. Additionally, the interleaved image-text approach, while effective for most tasks, may encounter difficulties in maintaining precise temporal alignment in very long video sequences. However, given its strengths in multimodal integration and extended context comprehension, LLaVA-Next-Video was selected as a benchmark model for this study, suited to VideoQA applications that demand a thorough understanding of video event sequences and interaction patterns.

\subsection{Generative Pre-trained Transformer $4$ Omni (GPT-$4o$)}
GPT-$4o$ \cite{gpt4o2024}, is a SOTA transformer-based model designed to process text, audio, and video in a unified architecture. Renowned for its advanced multimodal capabilities, GPT-$4o$ excels in VideoQA by integrating and interpreting diverse data types with high precision and efficiency. The model incorporates key characteristics to enhance performance and scalability including:
\begin{itemize}
\item \textbf{Sparse attention mechanisms:} Focus computational resources on the most relevant data segments, reducing processing overhead and improving efficiency.
\item \textbf{Layer pruning} Streamlines the architecture by removing redundant layers, enabling faster real-time processing.
\item \textbf{Mixed precision training:} Balances $16$-bit and $32$-bit operations to optimize memory usage while maintaining accuracy.
\end{itemize}
These features collectively enable GPT-$4o$ to handle complex multimodal tasks with high efficiency and adaptability. By seamlessly integrating inputs from text, audio, and video, the model supports cross-modal interpretation and alignment, making it ideal for scenarios requiring a nuanced understanding of multimodal data (See Figure \ref{fig05} (a) as an example). GPT-$4o$ was trained on proprietary datasets spanning text, audio, image, and video domains, ensuring broad coverage and adaptability. Its performance on VideoQA tasks was evaluated on benchmarks such as ActivityNet \cite{7298698} and EgoSchema \cite{egoschema}, achieving scores of 61.9\% and 72.2\%, respectively.
GPT-$4o$ demonstrates significant strengths, including advanced multimodal integration that enables coherent interpretation of diverse input types and optimized real-time performance through efficient computational strategies like sparse attention and layer pruning. However, it has some limitations, such as high computational demands for large-scale tasks and sensitivity to data quality and diversity, with challenges in handling certain audio-based inputs \cite{gpt4o2024}. Despite these drawbacks, GPT-$4o$'s innovative design and robust architecture make it a benchmark model for VideoQA. Its multimodal capabilities and ability to handle complex queries efficiently have positioned it as a critical tool for advancing multimodal AI research. Additionally, its capacity to adaptively improve and correct errors through internet-crawled data further highlights its potential as a leading model in the field \cite{li2024llava, liu2024llavanext, li2024llavanext-strong, geminiteam2024geminifamilyhighlycapable}.

\subsection{Gemini \textbf{$1.5$} Pro }

\begin{figure}[t]
\centering{\includegraphics[width=0.90\linewidth]{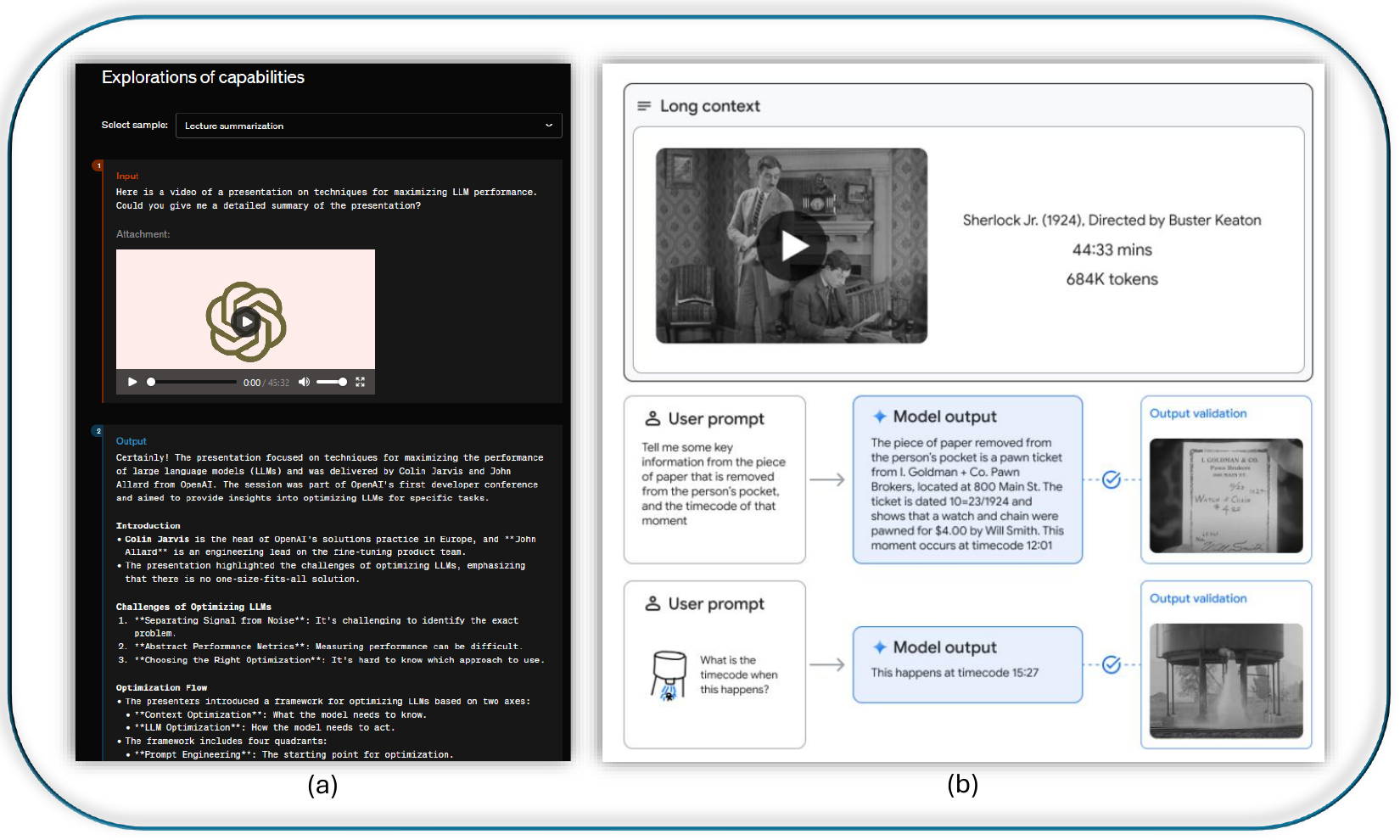}}
\caption{(a) Shows an extensive summary of a $45mins$ lecture provided by GPT-$4o$, it includes subjection for each topic discussed in the video \cite{gpt4o2024}, (b) represents the Gemini Pro VideoQA capabilities the movie Sherlock Jr. of $44mins$ is passed. The superiority of the model to answer questions as well as recognize hand-drawn images and justify the answer by providing the exact frame as validation figure referenced from \cite{geminiteam2024geminifamilyhighlycapable}.}
\label{fig05}
\end{figure}

Gemini $1.5$ Pro \cite{geminiteam2024geminifamilyhighlycapable} is a model designed to handle diverse data inputs, including text, audio, and video, within extensive spatiotemporal contexts. Its architecture and training capabilities enable it to excel in complex tasks such as VideoQA with a particular emphasis on precision and scalability across extremely lengthy data sequences (See Figure \ref{fig05} (b) as an example). Below are the standout features of Gemini $1.5$ Pro, which enable it to excel in long-form and complex multimodal tasks.

\begin{itemize}
\item \textbf{Needle-in-a-Haystack capability}: Gemini $1.5$ Pro \cite{geminiteam2024geminifamilyhighlycapable} is notable for its ability to locate specific details within large-scale contexts. This feature allows it to retrieve accurate information even when embedded in sequences as long as $10$ million tokens across multiple modalities.
\item \textbf{ 1H-VideoQA}: The model can recall embedded information from extended video content, exemplified by its performance on datasets containing up to $10.5$ hours of video. This capability underlines its robust memory management and retrieval performance, vital for long-form video and multimodal content analysis.
\item \textbf{High precision in retrieval Tasks}: With a recall rate of over $99.7\%$ \cite{geminiteam2024geminifamilyhighlycapable}, Gemini $1.5$ Pro excels in accurately retrieving specific content, even within extensive data sequences, which is essential for tasks requiring high levels of detail and precision.
\end{itemize}

The architecture of Gemini $1.5$ Pro incorporates several key elements optimized for multimodal processing and extensive context retention as summarized below.
\begin{itemize}
    \item \textbf{Multimodal input integration}: Gemini $1.5$ Pro processes data from various sources, including text, audio, and video, through a unified architecture that allows simultaneous interpretation and alignment across modalities.
    \item \textbf{Extended contextual memory}: The model is designed to handle very large contexts, leveraging an advanced memory management system that retains relevant details over prolonged sequences. This feature enhances its capability to track events across extended video or audio content without information loss.
    \item \textbf{Sparse attention mechanisms}: Sparse attention enables the model to focus selectively on pertinent segments of data, which reduces computational load while preserving performance in handling high-density video and audio data streams.
    \item \textbf{Layer pruning and mixed precision training}: To optimize processing efficiency, the architecture incorporates layer pruning and mixed precision training, balancing speed and performance without sacrificing accuracy in long-duration data tasks.
\end{itemize}

Gemini $1.5$ Pro demonstrates several key strengths that make it a powerful tool for multimodal tasks. Its versatility allows seamless interpretation and alignment of data across text, audio, and video inputs, enabling the model to handle complex tasks like VideoQA with high accuracy. Furthermore, the model excels in extended contextual retrieval, processing sequences of up to $10$ million tokens. This capability is particularly beneficial for applications that require detailed recall over extensive data streams, such as long-form video analysis. Additionally, resource efficiency is achieved through the incorporation of sparse attention mechanisms, layer pruning, and mixed precision training. These optimizations strike a balance between computational demands and performance, making the model suitable for both real-time and large-scale applications.
However, Gemini $1.5$ Pro also faces certain limitations. High computational demand remains a challenge for large-scale tasks, particularly when handling extensive video and audio modalities. Despite its efficiency strategies, processing such voluminous data requires substantial computational resources. Moreover, the model’s dependency on data quality and diversity can impact its performance. In scenarios requiring understanding across various contexts and modalities, a lack of diverse or high-quality training data may hinder its effectiveness.

Gemini $1.5$ Pro was included in this study for its unique capabilities in VideoQA, supported by its strong retrieval accuracy and capacity to handle extended contexts. Its robust multimodal processing abilities make it suitable for tasks involving complex question-answering and long-form content retrieval, positioning it as a benchmark model for high-precision VideoQA and other demanding multimodal applications. 
Table \ref{table02} presents a comparative overview of the SOTA VideoQA models evaluated in this study. It outlines their key features, including support for multimodal inputs, long video processing, fine-tuning capabilities, and primary use cases, offering a clear view of their respective strengths and differences.

\begin{table}[t]
\caption{Comparison of key features and capabilities of SOTA VideoQA models used in this study.}
\label{table02}
\resizebox{\textwidth}{!}{%
\begin{tabular}{l|ccccc}
\toprule \hline
\rowcolor[HTML]{EFEFEF}
\multicolumn{1}{c|}{\cellcolor[HTML]{EFEFEF}} &
  \multicolumn{5}{c}{\cellcolor[HTML]{EFEFEF}\textbf{State-of-the-art VideoQA Models used in this Study}} \\ \cline{2-6} 
\rowcolor[HTML]{EFEFEF}
\multicolumn{1}{c|}{\multirow{-2}{*}{\cellcolor[HTML]{EFEFEF}\textbf{Features}}} &
  \multicolumn{1}{c|}{\cellcolor[HTML]{EFEFEF}\textbf{LLAVA-NeXT-7B}} &
  \multicolumn{1}{c|}{\cellcolor[HTML]{EFEFEF}\textbf{InternVL}} &
  \multicolumn{1}{c|}{\cellcolor[HTML]{EFEFEF}\textbf{VideoLLaMA-$2$}} &
  \multicolumn{1}{c|}{\cellcolor[HTML]{EFEFEF}\textbf{GPT-$4o$}} &
  \textbf{Gemini-$1.5$ Pro} \\ \hline \midrule
\rowcolor[HTML]{FFFFFF}
\cellcolor[HTML]{EFEFEF}\textbf{Open Source} &
  \multicolumn{1}{c|}{\textcolor{blue}{\ding{51}}} &
  \multicolumn{1}{c|}{\textcolor{blue}{\ding{51}}} &
  \multicolumn{1}{c|}{\textcolor{blue}{\ding{51}}} &
  \multicolumn{1}{c|}{\textcolor{red}{\ding{55}} } &
  \textcolor{red}{\ding{55}} \\ \hline
\rowcolor[HTML]{FFFFFF}
\cellcolor[HTML]{EFEFEF}\textbf{API Availability} &
  \multicolumn{1}{c|}{\textcolor{blue}{\ding{51}}} &
  \multicolumn{1}{c|}{\textcolor{blue}{\ding{51}}} &
  \multicolumn{1}{c|}{\textcolor{red}{\ding{55}}} &
  \multicolumn{1}{c|}{\textcolor{blue}{\ding{51}}} &
  \textcolor{blue}{\ding{51}} \\ \hline
\rowcolor[HTML]{FFFFFF}
\cellcolor[HTML]{EFEFEF}\textbf{Fine Tuning} &
  \multicolumn{1}{c|}{\textcolor{blue}{\ding{51}}} &
  \multicolumn{1}{c|}{\textcolor{blue}{\ding{51}}} &
  \multicolumn{1}{c|}{\textcolor{blue}{\ding{51}}} &
  \multicolumn{1}{c|}{\textcolor{blue}{\ding{51}}} &
  \textcolor{blue}{\ding{51}} \\ \hline
\rowcolor[HTML]{FFFFFF}
\cellcolor[HTML]{EFEFEF}\textbf{Long Video Support} &
  \multicolumn{1}{c|}{\textcolor{blue}{\ding{51}}} &
  \multicolumn{1}{c|}{\textcolor{red}{\ding{55}}} &
  \multicolumn{1}{c|}{\textcolor{red}{\ding{55}}} &
  \multicolumn{1}{c|}{\textcolor{blue}{\ding{51}}} &
  \textcolor{blue}{\ding{51}} \\ \hline
\rowcolor[HTML]{FFFFFF}
\cellcolor[HTML]{EFEFEF}\textbf{Supports Complex Tasks} &
  \multicolumn{1}{c|}{\textcolor{blue}{\ding{51}}} &
  \multicolumn{1}{c|}{\textcolor{blue}{\ding{51}}} &
  \multicolumn{1}{c|}{\textcolor{blue}{\ding{51}}} &
  \multicolumn{1}{c|}{\textcolor{blue}{\ding{51}}} &
  \textcolor{blue}{\ding{51}} \\ \hline
\rowcolor[HTML]{FFFFFF}
\cellcolor[HTML]{EFEFEF}\textbf{Temporal Processing} &
  \multicolumn{1}{c|}{\textcolor{blue}{\ding{51}}} &
  \multicolumn{1}{c|}{\textcolor{blue}{\ding{51}}} &
  \multicolumn{1}{c|}{\textcolor{blue}{\ding{51}}} &
  \multicolumn{1}{c|}{\textcolor{blue}{\ding{51}}} &
  \textcolor{blue}{\ding{51}} \\ \hline
\rowcolor[HTML]{FFFFFF}
\cellcolor[HTML]{EFEFEF}\textbf{Spatial Reasoning} &
  \multicolumn{1}{c|}{\textcolor{blue}{\ding{51}}} &
  \multicolumn{1}{c|}{\textcolor{blue}{\ding{51}}} &
  \multicolumn{1}{c|}{\textcolor{blue}{\ding{51}}} &
  \multicolumn{1}{c|}{\textcolor{blue}{\ding{51}}} &
  \textcolor{blue}{\ding{51}} \\ \hline
\rowcolor[HTML]{FFFFFF}
\cellcolor[HTML]{EFEFEF}\textbf{Primary Use Case} &
  \multicolumn{1}{c|}{\begin{tabular}[c]{@{}c@{}}Long video QA,\\ visual QA\end{tabular}} &
  \multicolumn{1}{c|}{\begin{tabular}[c]{@{}c@{}}Video and visual\\ QA\end{tabular}} &
  \multicolumn{1}{c|}{\begin{tabular}[c]{@{}c@{}}Video-text\\ alignment\end{tabular}} &
  \multicolumn{1}{c|}{\begin{tabular}[c]{@{}c@{}}MultiModal\\ Agent\end{tabular}} &
  \begin{tabular}[c]{@{}c@{}}MultiModal\\ Agent\end{tabular} \\ \hline
\rowcolor[HTML]{FFFFFF}
\cellcolor[HTML]{EFEFEF}\textbf{Multimodality} &
  \multicolumn{1}{c|}{Text, Video} &
  \multicolumn{1}{c|}{Text, Video} &
  \multicolumn{1}{c|}{Text, Video, Audio} &
  \multicolumn{1}{c|}{Text, Video, Audio} &
  Text, Video, Audio \\ \hline
\rowcolor[HTML]{FFFFFF}
\cellcolor[HTML]{EFEFEF}\textbf{Architecture} &
  \multicolumn{1}{c|}{\begin{tabular}[c]{@{}c@{}}ViT-L/14 \\ Vicuna \end{tabular}} &
  \multicolumn{1}{c|}{\begin{tabular}[c]{@{}c@{}}ViT-MLP-LLM \\ InternViT-6B\end{tabular}} &
  \multicolumn{1}{c|}{\begin{tabular}[c]{@{}c@{}}CLIP ViT-L/14 \\ with Mistral-Instruct\end{tabular}} &
  \multicolumn{1}{c|}{Sparse MoE Transformer} &
  Sparse MoE Transformer \\ \hline \bottomrule

\end{tabular}%
}
\end{table}


\section{Evaluation Framework}
\label{sect_05}

\begin{figure}[t]
\centering{\includegraphics[width=0.90\linewidth]{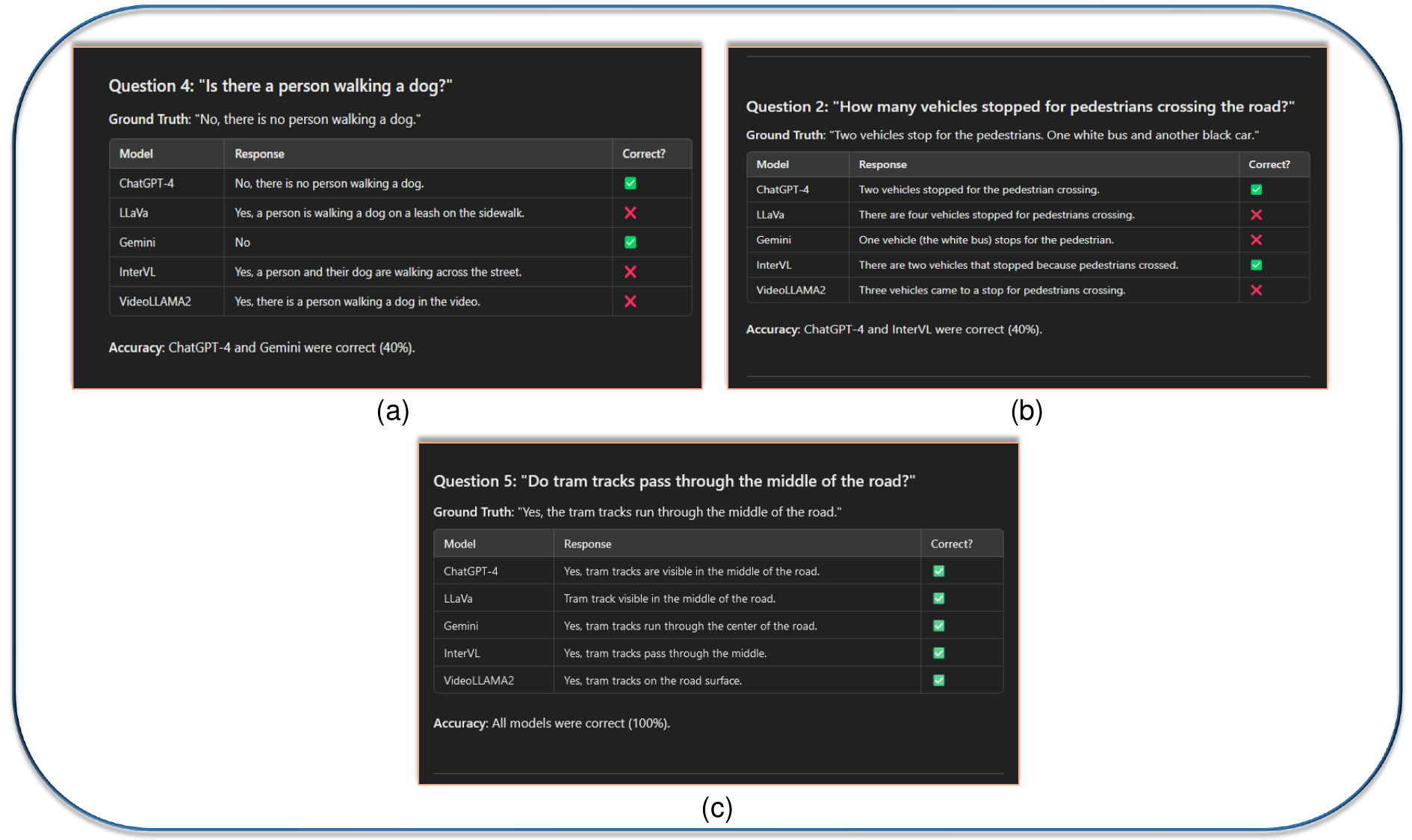}}
\caption{An overview of the output tables based on the prompt. (a) are the scores generated on  questions taken from sequence $1$, (b) are on the questions taken from the synthetic video sequence, and (c) are based on  questions based on the real-world sequence .}
\label{fig01_section5}
\end{figure}
This evaluation framework combines principles from two influential studies: the work by Marcu et al. \cite{marcu2024lingoqavisualquestionanswering}, and Liao et al. \cite{liao2024alignaggregatecompositionalreasoning}, which explores the use of language models as objective evaluators, and the approach that emphasizes compositional reasoning with consistency metrics in VideoQA. Thus integrating these methodologies enables a structured and balanced assessment framework, leveraging the strengths of both studies.

\subsection{Language Models as Evaluators, Compositional Reasoning, and Consistency Metrics} 
Marcu et al.\cite{marcu2024lingoqavisualquestionanswering}, introduces the use of LLMs for evaluating VideoQA, and Liao et al. \cite{liao2024alignaggregatecompositionalreasoning}, focuses on compositional reasoning through consistency metrics. These approaches combine to form a structured assessment framework.
Marcu et al. \cite{marcu2024lingoqavisualquestionanswering} proposed using language models like Lingo-Judge as evaluators to score responses based on criteria such as relevance, accuracy, and consistency. The method aligned model-generated answers with ground truth, showing that language models can objectively evaluate VideoQA performance without direct access to visual content. In this framework, GPT-$4o$ serves as the primary evaluator, assessing responses based on these criteria.
Liao et al. 
\cite{liao2024alignaggregatecompositionalreasoning} introduced the Video Alignment and Answer Aggregation (VA$3$) framework, which evaluates compositional reasoning using consistency metrics. These metrics measure how well models handle related questions within the same context. This framework integrates the cP metric into GPT-$4o$’s evaluation process to assess the consistency of model responses across similar queries.
By combining quality evaluation and consistency analysis, this framework provides a comprehensive assessment of model performance across various question types and levels of complexity.

 \subsection{ GPT-$4o$ as a Baseline Evaluator for Traffic VideoQA}

GPT-$4o$ was prompted to act as an evaluator by assessing the accuracy of model-generated responses against predefined ground truth answers. To ensure impartiality, GPT-$4o$ was not provided with the actual video footage but only with the ground truth and the responses of the VideoQA models. This approach focused on semantic accuracy, capturing whether the model responses conveyed the same meaning as the ground truth.
A score of $1$ was assigned if the response was semantically consistent with the ground truth, and $0$ otherwise. This scoring method ensured that the evaluation remained objective and focused solely on aligning the ground truth and the model's output (refer Figure \ref{fig01_section5}). By leveraging GPT-$4o$'s strong natural language understanding, the evaluation effectively measured the semantic correctness of responses, which is crucial for the reliability of VideoQA models in traffic analysis scenarios.

\subsubsection{ Prompt Design for Evaluation }
An example prompt provided for GPT is as follows:

\begin{center}
\textit{
Create a table to compare the performance of VideoQA models on three different question sets related to traffic monitoring conditions. For each question set:
\begin{enumerate}
    \item Provide the \textbf{ground truth} for the question.
    \item List the \textbf{model responses}.
    \item Indicate whether the responses were \textbf{correct} (score as 1 for semantically correct and 0 for completely wrong; no partial grading).
    \item Calculate the \textbf{accuracy percentage} for the models as:
    \[
    \text{Accuracy (\%)} = \frac{\text{Number of correct responses}}{\text{Total number of responses}} \times 100
    \]
    \item Calculate the \textbf{consistency precision} (\( c_P \)) for the models as:
    \[
    c_P = \frac{\text{Total number of questions answered correctly with all questions in the category also correct}}{\text{Total number of questions with all questions answered correctly}}
    \]
\end{enumerate}
}
\end{center}

The prompt outlines the task of comparing the performance of VideoQA models on three sets of questions related to traffic monitoring scenarios. For each question set, the evaluation involves providing the ground truth and the responses from the models. The responses were scored with no partial grading. The evaluation requires calculating two metrics: the \textbf{accuracy percentage}, which measures the proportion of correct responses out of the total responses, and the \textbf{cP}, which assesses how consistently a model answers all questions correctly within a category. The goal is to analyze the semantic alignment between the model responses and the ground truth, enabling a standardized and objective performance comparison of the VideoQA models in the context of traffic analysis.

\section{Experiments}
\label{experiment}
Following the evaluation of frameworks outlined (refer Section \ref{sect_05}), we present the findings from our experiments conducted on traffic video sequences (refer Section \ref{method-of-study}). The experiments were performed on five SOTA models (refer Section \ref{sect_04}) where the accuracy and performance scores for each traffic video sequence, encompassing both real-world and synthetic datasets were recorded. These sequences were evaluated over a range of question complexities, categorized as easy, moderate, and complex (refer appendix \ref{appendix}). Additionally, we provide a summary of the overall results of the model's performance across various scenarios.

\begin{figure}[t]
\centering{\includegraphics[width=0.90\linewidth]{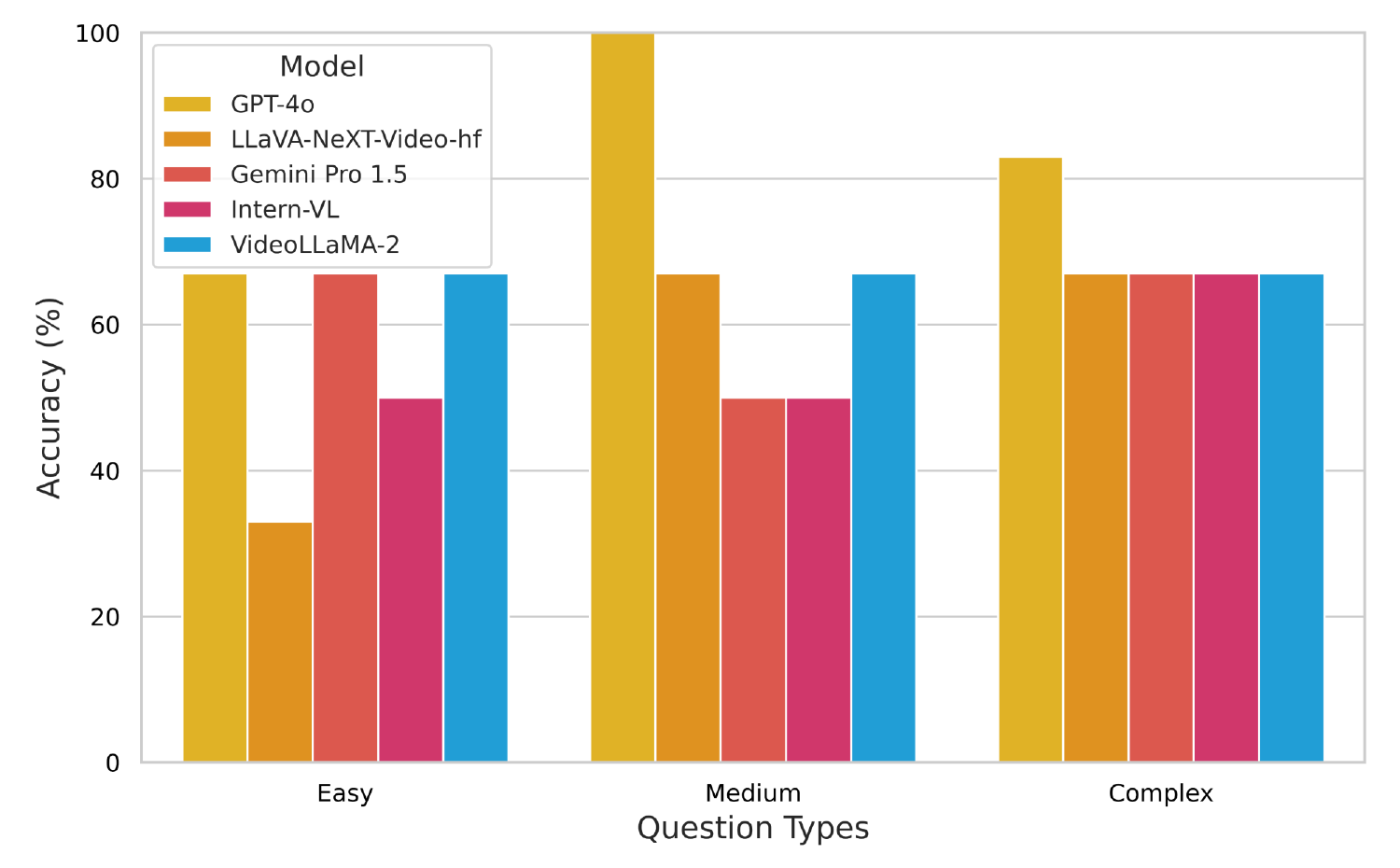}}
\caption{Average accuracy across question types for video sequence 1}
\label{fig06_1}
\end{figure}

\begin{table}[t]
\centering 
\caption{Average accuracy of VideoQA models on video sequence 1}
\label{table_03}

\begin{tabular*}{0.8\textwidth}{ @{\extracolsep{\fill}}
>{\centering\arraybackslash\columncolor[HTML]{EFEFEF}}l |
>{\centering\arraybackslash\columncolor[HTML]{FFFFFF}}c 
>{\centering\arraybackslash\columncolor[HTML]{FFFFFF}}c 
>{\centering\arraybackslash\columncolor[HTML]{FFFFFF}}c
}
\toprule
\hline
\multicolumn{1}{c|}{\cellcolor[HTML]{EFEFEF}} & \multicolumn{3}{c}{\cellcolor[HTML]{EFEFEF}\textbf{SOTA VideoQA Models Performance}}                         \\ \cline{2-4} 
\multicolumn{1}{c|}{\multirow{-2}{*}{\cellcolor[HTML]{EFEFEF}\textbf{VideoQA Models}}} &
  \multicolumn{1}{c|}{\cellcolor[HTML]{EFEFEF}\textbf{Easy Question}} &
  \multicolumn{1}{c|}{\cellcolor[HTML]{EFEFEF}\textbf{\centering Moderate Type}} &
  \cellcolor[HTML]{EFEFEF}\textbf{\centering Complex Questions } \\ \hline \midrule
\multicolumn{1}{c|}{\cellcolor[HTML]{EFEFEF}\textbf{GPT-4o}} & \multicolumn{1}{c|}{\cellcolor[HTML]{FFFFFF}67\%} & \multicolumn{1}{c|}{\cellcolor[HTML]{FFFFFF}100\%} & 83\% \\ \hline
\multicolumn{1}{c|}{\cellcolor[HTML]{EFEFEF}\textbf{LLaVA-NeXT-Video-7B-hf}}
                 & \multicolumn{1}{c|}{\cellcolor[HTML]{FFFFFF}33\%} & \multicolumn{1}{c|}{\cellcolor[HTML]{FFFFFF}67\%}  & 67\% \\ \hline 
\multicolumn{1}{c|}{\cellcolor[HTML]{EFEFEF}\textbf{Gemini Pro 1.5}}& \multicolumn{1}{c|}{\cellcolor[HTML]{FFFFFF}67\%} & \multicolumn{1}{c|}{\cellcolor[HTML]{FFFFFF}50\%}  & 67\% \\ \hline
\multicolumn{1}{c|}{\cellcolor[HTML]{EFEFEF}\textbf{InternVL}}                             & \multicolumn{1}{c|}{\cellcolor[HTML]{FFFFFF}50\%} & \multicolumn{1}{c|}{\cellcolor[HTML]{FFFFFF}50\%}  & 67\% \\ \hline
\multicolumn{1}{c|}{\cellcolor[HTML]{EFEFEF}\textbf{VideoLLaMA-2}}                         & \multicolumn{1}{c|}{\cellcolor[HTML]{FFFFFF}67\%} & \multicolumn{1}{c|}{\cellcolor[HTML]{FFFFFF}67\%}  & 67\% \\ \hline \bottomrule
\end{tabular*}

\end{table}

\subsection{Real-World Traffic Video Sequence $1$}
Sequence 1, which featured the cyclist (refer Section \ref{method-of-study} and Figure \ref{fig03} (a)), involved questions focused on the cyclist (the primary subject) and their surroundings (refer Appendix \ref{appendix}). For straightforward questions requiring basic observations, such as whether the cyclist followed the bike lane or the presence of pedestrians, ChatGPT-$4o$, Gemini, and VideoLLAMA2 achieved $67\%$ (refer Table \ref{table_03} and Figure \ref{fig06_1}) accuracy by correctly identifying details like the cyclist not adhering to the bike lane. Interestingly, both Gemini and GPT-$4o$ failed to answer a question regarding the presence of a dog in the video (refer to Table \ref{table_03}). LLaVA-NeXT-Video-$7B$-hf struggled the most, achieving only $33\%$ accuracy, while InternVL performed moderately with $50\%$.

For questions of moderate difficulty, which required analyzing interactions, GPT-$4o$ excelled, achieving $100\%$ accuracy by consistently identifying details such as the number of vehicles in motion. LLaVA-NeXT-Video-7B-hf and VideoLLAMA2 followed with $67\%$ accuracy, performing well in certain aspects but faltering in others. Gemini Pro $1.5$ and InternVL trailed with only $50\%$ accuracy, displaying noticeable inconsistencies in their responses.
In the case of hard questions, which demanded a deeper understanding of traffic safety and cyclist behaviour, GPT-$4o$ performed relatively well, achieving $83\%$ accuracy. It provided detailed and accurate explanations regarding how the cyclist’s actions impacted traffic safety, suggested improvements for bicycle-vehicle segregation, and correctly identified that the cyclist did not slow down. The other models demonstrated similar performance, averaging $67\%$ accuracy overall. However, they exhibited significant weaknesses in consistency and precision, often providing answers that were either inconsistent or lacked the required level of detail and clarity.

\begin{figure}[t]
\centering{\includegraphics[width=0.90\linewidth]{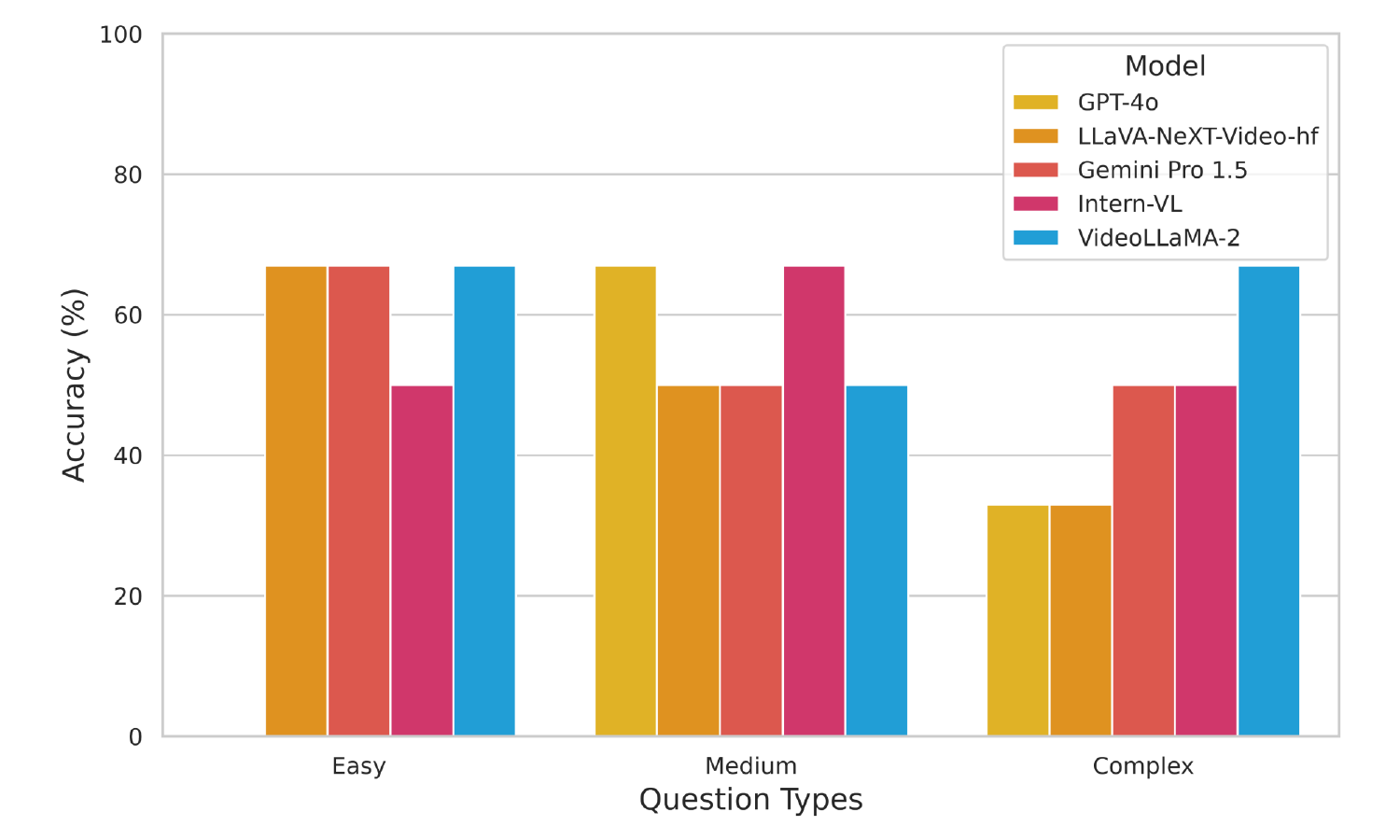}}
\caption{Average accuracy across question types for video sequence 2}
\label{fig06_2}
\end{figure}

\begin{table}[t]
\centering 
\caption{Average accuracy of VideoQA models on video sequence 2}
\label{table_0602}

\begin{tabular*}{0.8\textwidth}{ @{\extracolsep{\fill}}
>{\centering\arraybackslash\columncolor[HTML]{EFEFEF}}l |
>{\centering\arraybackslash\columncolor[HTML]{FFFFFF}}c 
>{\centering\arraybackslash\columncolor[HTML]{FFFFFF}}c 
>{\centering\arraybackslash\columncolor[HTML]{FFFFFF}}c
}
\toprule
\hline
\multicolumn{1}{c|}{\cellcolor[HTML]{EFEFEF}} & \multicolumn{3}{c}{\cellcolor[HTML]{EFEFEF}\textbf{SOTA VideoQA Models Performance}}                         \\ \cline{2-4} 
\multicolumn{1}{c|}{\multirow{-2}{*}{\cellcolor[HTML]{EFEFEF}\textbf{VideoQA Models}}} &
  \multicolumn{1}{c|}{\cellcolor[HTML]{EFEFEF}\textbf{Easy Question}} &
  \multicolumn{1}{c|}{\cellcolor[HTML]{EFEFEF}\textbf{\centering Moderate Type}} &
  \cellcolor[HTML]{EFEFEF}\textbf{\centering Complex Questions } \\ \hline \midrule
\multicolumn{1}{c|}{\cellcolor[HTML]{EFEFEF}\textbf{GPT-4o}}                              & \multicolumn{1}{c|}{\cellcolor[HTML]{FFFFFF}0\%} & \multicolumn{1}{c|}{\cellcolor[HTML]{FFFFFF}67\%} & 33\% \\ \hline
\multicolumn{1}{c|}{\cellcolor[HTML]{EFEFEF}\textbf{LLaVA-NeXT-Video-7B-hf}}              & \multicolumn{1}{c|}{\cellcolor[HTML]{FFFFFF}67\%} & \multicolumn{1}{c|}{\cellcolor[HTML]{FFFFFF}50\%}  & 33\% \\ \hline
\multicolumn{1}{c|}{\cellcolor[HTML]{EFEFEF}\textbf{Gemini Pro 1.5}}                      & \multicolumn{1}{c|}{\cellcolor[HTML]{FFFFFF}67\%} & \multicolumn{1}{c|}{\cellcolor[HTML]{FFFFFF}50\%}  & 50\% \\ \hline
\multicolumn{1}{c|}{\cellcolor[HTML]{EFEFEF}\textbf{InternVL}}                              & \multicolumn{1}{c|}{\cellcolor[HTML]{FFFFFF}50\%} & \multicolumn{1}{c|}{\cellcolor[HTML]{FFFFFF}67\%}  & 50\% \\ \hline
\multicolumn{1}{c|}{\cellcolor[HTML]{EFEFEF}\textbf{VideoLLaMA-2}}                         & \multicolumn{1}{c|}{\cellcolor[HTML]{FFFFFF}67\%} & \multicolumn{1}{c|}{\cellcolor[HTML]{FFFFFF}50\%}  & 67\% \\ \hline \bottomrule
\end{tabular*}
\end{table}
\subsection{Real-World Traffic Video Sequence $2$}
Sequence $2$, which involved cars turning onto the bike lane (refer Section \ref{wrong_turn} and Figure \ref{fig03} (b)). In terms of the easy questions (refer Appendix \ref{appendix}) LLaVA-NeXT-Video-$7B$-hf, Gemini Pro $1.5$, and VideoLLaMA-$2$ performed the best with a $67\%$ accuracy (refer Figure \ref{fig06_2} and Table \ref{table_0602}), correctly identifying features such as lamp posts, trucks, although making errors in counts. ChatGPT-$4o$ performed poorly, providing highly inaccurate answers and frequently failing to correctly identify details like the number of lamp posts or the presence of trucks. InternVL achieved $50\%$ accuracy but exhibited inconsistencies in identifying bicycles and trucks. The models struggled with questions requiring precise counts or distinguishing between objects, with Gemini $1.5$ overestimating pedestrians and GPT-$4o$ missing visible bicycles. 

For moderately difficult questions, GPT-$4o$ and InternVL performed the best with 67\% accuracy, accurately identifying features such as the number of vehicles turning onto the bike lane and the tram tracks passing through the middle of the road. LLaVA-NeXT-Video-$7B$-hf, Gemini Pro $1.5$, and VideoLLAMA-$2$ followed with 50\% accuracy, often making errors on questions related to vehicle signalling and shared lane signage. GPT-$4o$ and InternVL  were more reliable in interpreting traffic dynamics, such as identifying whether vehicles signalled or the presence of manhole covers, whereas the other models often provided contradictory or unclear responses. For complex questions, VideoLLAMA2 performed the best with  $67\%$ accuracy, successfully identifying intricate patterns such as the color of the most common car and vehicle types. Gemini Pro $1.5$ and InternVL followed with $50\%$ accuracy, providing some correct details but struggling with contextual reasoning, such as explaining intersection design or identifying returning pedestrians. GPT-$4o$ and LLaVA-NeXT-Video-$7B$-hf scored the lowest, with $33\%$ accuracy, often failing to analyze scenarios requiring detailed interpretation, such as identifying vehicle types or explaining how intersection design impacts behavior. Across all levels of difficulty, the models demonstrated significant inconsistencies, frequently providing different outputs upon repeated iterations.

\begin{figure}[t]
\centering{\includegraphics[width=0.90\linewidth]{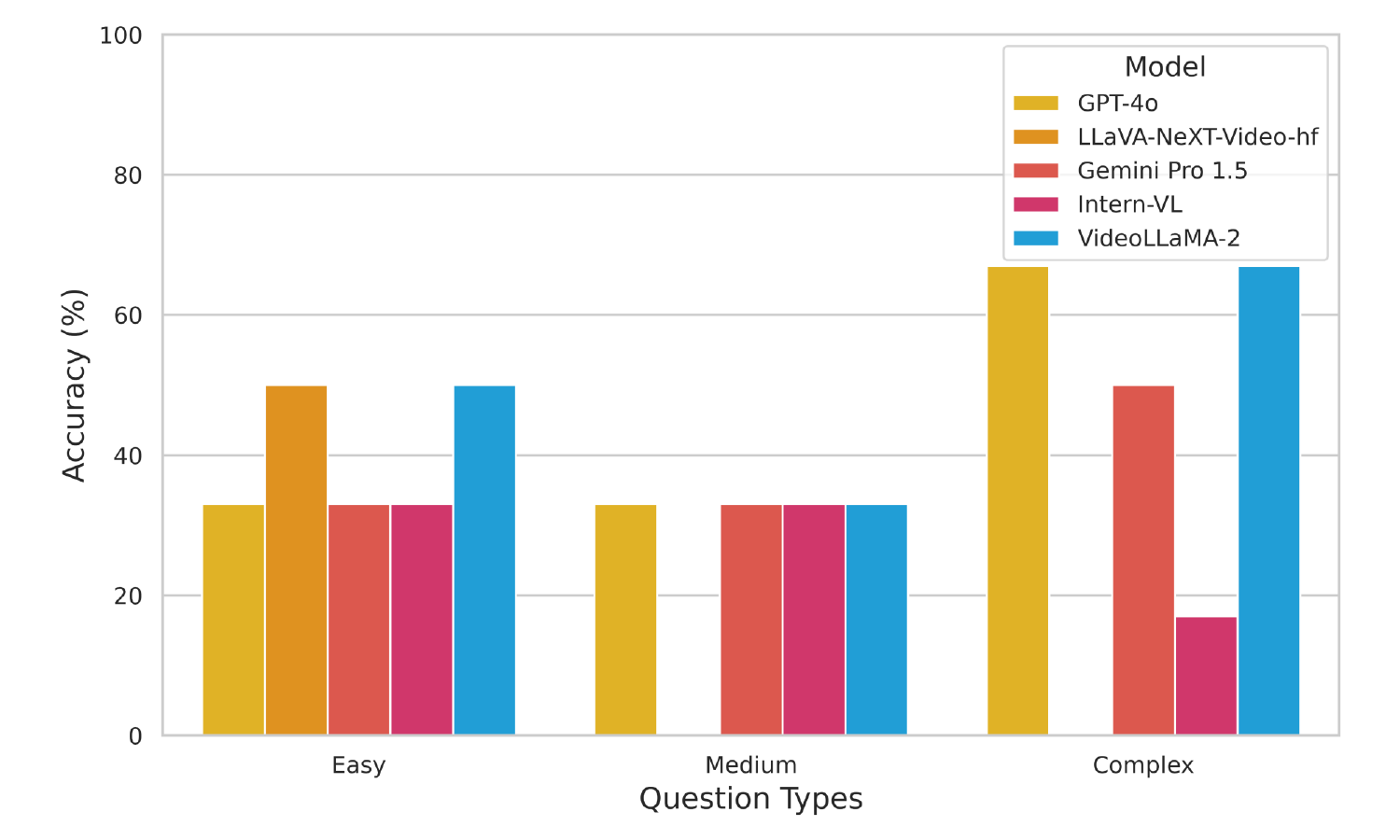}}
\caption{Average accuracy across question types for video sequence 3}
\label{fig06_3}
\end{figure}

\begin{table}[t]
\centering 
\caption{Average accuracy of VideoQA models on video sequence 3}
\label{tab:synthetic-table}

\begin{tabular*}{0.8\textwidth}{ @{\extracolsep{\fill}}
>{\centering\arraybackslash\columncolor[HTML]{EFEFEF}}l |
>{\centering\arraybackslash\columncolor[HTML]{FFFFFF}}c 
>{\centering\arraybackslash\columncolor[HTML]{FFFFFF}}c 
>{\centering\arraybackslash\columncolor[HTML]{FFFFFF}}c
}
\toprule
\hline
\multicolumn{1}{c|}{\cellcolor[HTML]{EFEFEF}} & \multicolumn{3}{c}{\cellcolor[HTML]{EFEFEF}\textbf{SOTA VideoQA Models Performance}}                         \\ \cline{2-4} 
\multicolumn{1}{c|}{\multirow{-2}{*}{\cellcolor[HTML]{EFEFEF}\textbf{VideoQA Models}}} &
  \multicolumn{1}{c|}{\cellcolor[HTML]{EFEFEF}\textbf{Easy Question}} &
  \multicolumn{1}{c|}{\cellcolor[HTML]{EFEFEF}\textbf{\centering Moderate Type}} &
  \cellcolor[HTML]{EFEFEF}\textbf{\centering Complex Questions } \\ \hline \midrule
\multicolumn{1}{c|}{\cellcolor[HTML]{EFEFEF}\textbf{GPT-4o}}                                & \multicolumn{1}{c|}{\cellcolor[HTML]{FFFFFF}33\%} & \multicolumn{1}{c|}{\cellcolor[HTML]{FFFFFF}33\%} & 67\% \\ \hline
\multicolumn{1}{c|}{\cellcolor[HTML]{EFEFEF}\textbf{LLaVA-NeXT-Video-7B-hf}}                & \multicolumn{1}{c|}{\cellcolor[HTML]{FFFFFF}50\%} & \multicolumn{1}{c|}{\cellcolor[HTML]{FFFFFF}0\%}  & 0\% \\ \hline
\multicolumn{1}{c|}{\cellcolor[HTML]{EFEFEF}\textbf{Gemini Pro 1.5}}                        & \multicolumn{1}{c|}{\cellcolor[HTML]{FFFFFF}33\%} & \multicolumn{1}{c|}{\cellcolor[HTML]{FFFFFF}33\%}  & 50\% \\ \hline
\multicolumn{1}{c|}{\cellcolor[HTML]{EFEFEF}\textbf{InternVL}}                              & \multicolumn{1}{c|}{\cellcolor[HTML]{FFFFFF}33\%} & \multicolumn{1}{c|}{\cellcolor[HTML]{FFFFFF}33\%}  & 17\% \\ \hline
\multicolumn{1}{c|}{\cellcolor[HTML]{EFEFEF}\textbf{VideoLLaMA-2}}                        & \multicolumn{1}{c|}{\cellcolor[HTML]{FFFFFF}50\%} & \multicolumn{1}{c|}{\cellcolor[HTML]{FFFFFF}33\%}  & 67\% \\ \hline \bottomrule
\end{tabular*}
\end{table}

\subsection{Synthetic Traffic Video Sequence $3$}
Sequence $3$ (refer Section \ref{synthetic}) involved a simulated intersection using CARLA (refer Figure \ref{fig03} (c)). In terms of straightforward easy questions (refer Appendix \ref{appendix}) focusing on simple observations, such as whether a pedestrian is using a crosswalk or the number of cars visible, LLaVA-NeXT-Video-7B-hf and VideoLLaMA-2 achieved the highest accuracy of $50\%$ (refer Figure \ref{fig06_3} and Table \ref{tab:synthetic-table}). ChatGPT-$4o$, Gemini Pro $1.5$, and InternVL lagged with $33\%$ accuracy, struggling with precise counts or nuanced details, such as the number of yellow cars or visible traffic lights. For moderate questions, which involved interpreting interactions such as the number of pedestrians crossing or vehicles stopping, all models performed poorly, scoring $33\%$ accuracy, except for LLaVA-NeXT-Video-$7B$-hf, which failed to provide a single correct answer. The models frequently miscounted pedestrians and vehicles or misunderstood directional movements. For instance, ChatGPT-$4o$ and Gemini Pro $1.5$ underestimated the number of pedestrians crossing and failed to correctly identify which vehicle arrived first at the intersection. For complex questions demanding deeper reasoning and analysis, ChatGPT-$4o$ demonstrated the highest accuracy $67\%$ accuracy, excelling in evaluating traffic light timing and identifying the number of vehicles in the intersection. VideoLLaMA-$2$ and Gemini Pro $1.5$ followed with moderate performance $50\%$, with Gemini Pro $1.5$ providing better reasoning about the impact of pedestrian behavior. LLaVA-NeXT-Video-$7B$-hf and InternVL struggled significantly in this category, delivering inaccurate and inconsistent results.

\begin{figure}[t]
\centering{\includegraphics[width=0.90\linewidth]{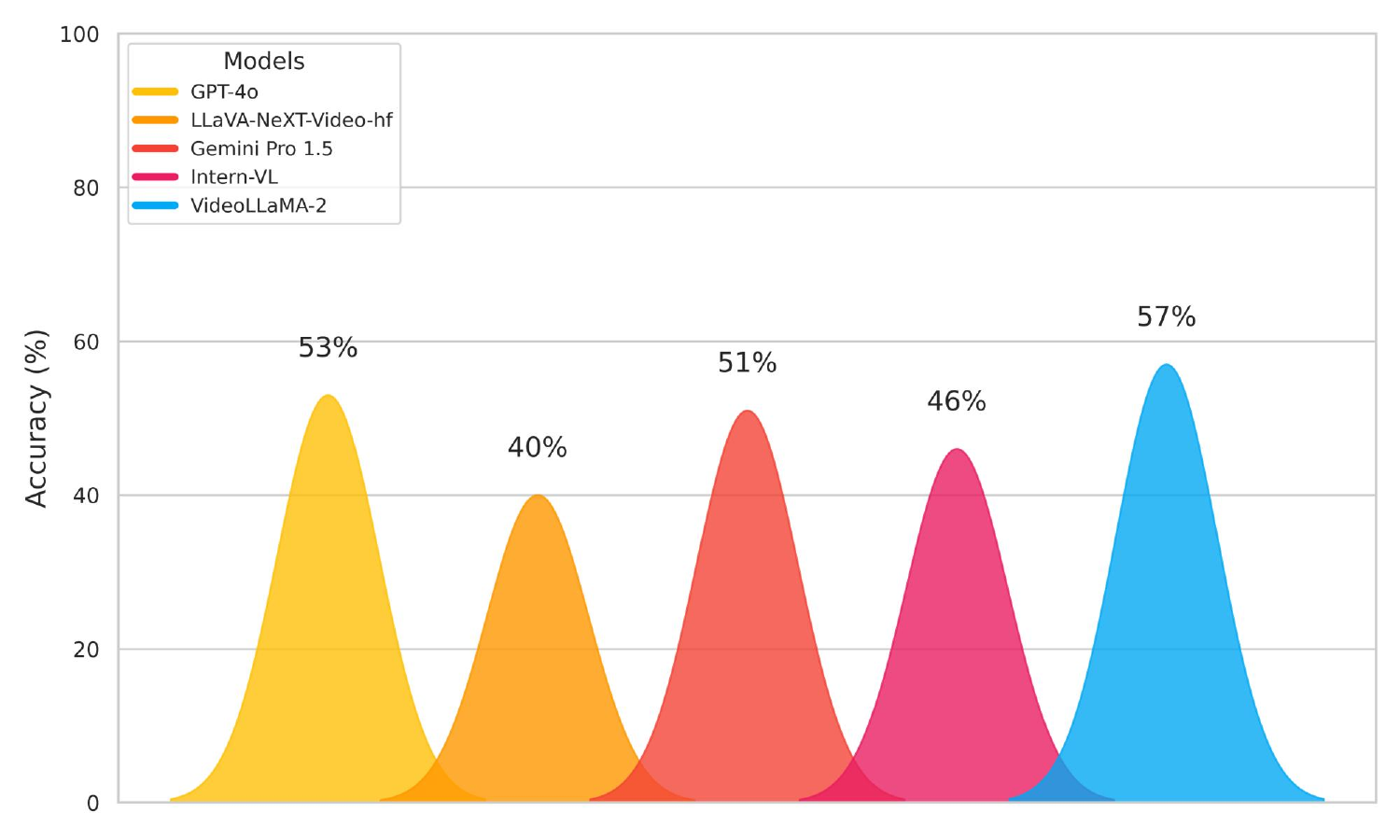}}
\caption{Average accuracy across all sequences and video types.}
\label{fig06_4}
\end{figure}

\begin{table}[]
\centering 
\caption{Average accuracy of VideoQA models across all video sequences}
\label{tab:overall}
\resizebox{0.6\textwidth}{!}{
\begin{tabular}{>{\columncolor[HTML]{EFEFEF}}l| >{\columncolor[HTML]{FFFFFF}}c}
\toprule \hline
\multicolumn{2}{c}{\cellcolor[HTML]{EFEFEF}\textbf{Overall Performance of SOTA VideoQA Models}} \\ \hline 
\multicolumn{1}{c|}{\cellcolor[HTML]{EFEFEF}\textbf{VideoQA Models}} & \cellcolor[HTML]{EFEFEF}\textbf{Average Consistency Precision} \\ \hline \midrule
\multicolumn{1}{c|}{\cellcolor[HTML]{EFEFEF}\textbf{GPT-4o}}                       & 53\%       \\ \hline
\multicolumn{1}{c|}{\cellcolor[HTML]{EFEFEF}\textbf{LLaVA-NeXT-Video-7B-hf}}        & 40\%       \\ \hline
\multicolumn{1}{c|}{\cellcolor[HTML]{EFEFEF}\textbf{Gemini Pro 1.5}}               & 51\%       \\ \hline
\multicolumn{1}{c|}{\cellcolor[HTML]{EFEFEF}\textbf{Intern-VL}}                    & 46\%       \\ \hline
\multicolumn{1}{c|}{\cellcolor[HTML]{EFEFEF}\textbf{VideoLLaMA-2}}                  & 57\%       \\ \hline \bottomrule
\end{tabular}%
}
\end{table}

\subsection{Overall Performance Across Models}
Across all sequences, VideoLLaMA-$2$  demonstrated strong performance in hard and moderate questions (refer Figure  \ref{fig06_4} and Table \ref{tab:overall}), particularly excelling in traffic safety analysis and complex reasoning tasks. However, it struggled with basic observations in synthetic sequences. InternVL showed inconsistent performance, performing better in straightforward observation tasks but struggling in reasoning-heavy scenarios.  Gemini Pro $1.5$ and GPT-$4o$ displayed moderate but inconsistent performance, with occasional strengths in scene interpretation but frequent errors in nuanced or context-heavy scenarios.LLaVA-NeXT-Video-$7B$-hf performed poorly with long descriptions without providing prompt answers for the questions.

These findings highlight both progress and ongoing limitations in VideoQA. While VideoLLaMA-2 shows potential in handling compositional reasoning and consistency, all models require further development in multi-object tracking, fine-grained attribute recognition, and temporal alignment. Addressing these areas would likely lead to improvements in model performance, particularly in real-world scenarios that demand high accuracy and a detailed understanding of video content.

Across all sequences, VideoLLaMA-2 demonstrated strong performance in hard and moderate questions, particularly excelling in traffic safety analysis and complex reasoning tasks. However, it struggled with basic observations in synthetic sequences. InternVL showed inconsistent performance, performing better in straightforward observation tasks but struggling in reasoning-heavy scenarios. Gemini Pro $1.5$ and GPT-$4o$ displayed moderate but inconsistent performance, with occasional strengths in scene interpretation but frequent errors in nuanced or context-heavy scenarios. LLaVA-NeXT-Video-$7$B-hf performed poorly, often providing long descriptions without delivering prompt answers to the questions. Overall, none of the models were consistent and precise with the answers generated. These findings highlight both progress and ongoing limitations in VideoQA. While VideoLLaMA-$2$ shows potential in handling compositional reasoning and consistency, all models require further development in multi-object tracking, fine-grained attribute recognition, and temporal alignment. Addressing these areas would likely lead to improvements in model performance, particularly in real-world scenarios that demand high accuracy and a detailed understanding of video content.

\section{Conclusion}

This study presents a detailed evaluation of the SOTA VideoQA models, including GPT-$4o$, LLaVA-NeXT-Video-7B-hf, Gemini Pro $1.5$, Intern-VL, and VideoLLaMA-$2$, focusing on their ability to handle varying levels of question complexity across diverse traffic video scenarios. VideoQA, as a multimodal task, demands the integration of spatial and temporal analysis, object tracking, and compositional reasoning, posing significant challenges for AI models. To address these, this study implemented an evaluation framework, using GPT-$4o$ as an impartial evaluator.

The results demonstrate that VideoLLaMA-$2$ stands out among the evaluated models, achieving an average accuracy of 57\%. It excelled in compositional reasoning and consistency across question types, especially in handling complex questions that required in-depth understanding of spatial-temporal relationships. GPT-$4o$  and Gemini Pro $1.5$, both scored 53\% and 51\% respectively following closely, showing strengths in handling simpler and moderately challenging questions but facing limitations in complex reasoning tasks. Intern-VL, with a score of 46\%,  showed moderate performance but struggled with multi-object detection. LLaVA-NeXT-Video-7B-hf, with the lowest score of 40\%, struggled particularly with quantitative tasks and complex interaction analysis, indicating room for substantial improvement. The key takeaways out of the experiments conducted in this study are:

\begin{itemize}
    \item \textbf{Strengths of the top models}: VideoLLaMA-$2$'s performance underscores the importance of architectures capable of compositional and counterfactual reasoning, which are critical for complex VideoQA tasks. The robustness of the model in handling various types of questions indicates its adaptability to real-world VideoQA applications, where the complexity of queries can vary significantly.
    
    \item \textbf{Areas for improvement}: Despite their strengths, even the top models showed limitations in tasks requiring high accuracy in multi-object detection and complex temporal reasoning. Improvements in object-tracking mechanisms, temporal alignment, and consistency in multi-object detection would enhance performance in complex VideoQA scenarios.

    \item \textbf{Implications of using GPT-$4o$ as an evaluator}: Leveraging GPT-$4o$ as a judge provided a consistent and objective framework for scoring responses across varied question types, particularly useful in a domain as complex as traffic monitoring. However, future studies could refine this evaluation by incorporating specialized checks for nuanced traffic-related queries, such as identifying incidents or tracking vehicle trajectories under varying weather and lighting conditions.
\end{itemize}

The insights from this study suggest that VideoQA models, with further development, could become powerful tools in intelligent transportation systems. The ability to automatically answer questions like \textit{“How many vehicles ran a red light?”} or \textit{“Was there a collision between 3:00 PM and 3:05 PM?”} would enhance traffic management, improve response times to incidents, and enable more efficient urban planning. However, current model limitations in multi-object tracking and temporal reasoning must be addressed to fully realize these benefits.
In conclusion,  this study contributes to the VideoQA field by combining objective evaluation with compositional metrics, identifying VideoLLaMA-2 as the most effective model while highlighting broader challenges. Future research should prioritize refining multi-object tracking and temporal reasoning to advance the applicability of VideoQA in real-world traffic monitoring scenarios, paving the way for smarter, more responsive transportation systems.

\newpage
\bibliographystyle{unsrt}  
\bibliography{references}  

\begin{thebibliography}{10}

\bibitem{zhong2022videoquestionansweringdatasets}
Yaoyao Zhong, Junbin Xiao, Wei Ji, Yicong Li, Weihong Deng, and Tat-Seng Chua.
\newblock Video question answering: Datasets, algorithms and challenges, 2022.

\bibitem{xiao2024videoqaerallmsempirical}
Junbin Xiao, Nanxin Huang, Hangyu Qin, Dongyang Li, Yicong Li, Fengbin Zhu, Zhulin Tao, Jianxing Yu, Liang Lin, Tat-Seng Chua, and Angela Yao.
\newblock Videoqa in the era of llms: An empirical study, 2024.

\bibitem{qasemi2023trafficdomainvideoquestionanswering}
Ehsan Qasemi, Jonathan~M. Francis, and Alessandro Oltramari.
\newblock Traffic-domain video question answering with automatic captioning, 2023.

\bibitem{dosovitskiy2017carlaopenurbandriving}
Alexey Dosovitskiy, German Ros, Felipe Codevilla, Antonio Lopez, and Vladlen Koltun.
\newblock Carla: An open urban driving simulator, 2017.

\bibitem{marcu2024lingoqavisualquestionanswering}
Ana-Maria Marcu, Long Chen, Jan Hünermann, Alice Karnsund, Benoit Hanotte, Prajwal Chidananda, Saurabh Nair, Vijay Badrinarayanan, Alex Kendall, Jamie Shotton, Elahe Arani, and Oleg Sinavski.
\newblock Lingoqa: Visual question answering for autonomous driving, 2024.

\bibitem{liao2024alignaggregatecompositionalreasoning}
Zhaohe Liao, Jiangtong Li, Li~Niu, and Liqing Zhang.
\newblock Align and aggregate: Compositional reasoning with video alignment and answer aggregation for video question-answering, 2024.

\bibitem{gpt4o2024}
{OpenAI}.
\newblock {GPT-4o: System Card}.
\newblock Available at: \url{https://cdn.openai.com/gpt-4o-system-card.pdf}, 2024.
\newblock \url{https://cdn.openai.com/gpt-4o-system-card.pdf}.

\bibitem{10582849}
Jie Nie, Xin Wang, Runze Hou, Guohao Li, Hong Chen, and Wenwu Zhu.
\newblock Dynamic spatio-temporal graph reasoning for videoqa with self-supervised event recognition.
\newblock {\em IEEE Transactions on Image Processing}, 33:4145--4158, 2024.

\bibitem{gan2016recognizing}
Chuang Gan, Yi~Yang, Linchao Zhu, Deli Zhao, and Yueting Zhuang.
\newblock Recognizing an action using its name: A knowledge-based approach.
\newblock {\em International Journal of Computer Vision}, 120:61--77, 2016.

\bibitem{zhu2019video}
Linchao Zhu.
\newblock {\em Video representation learning with deep neural networks}.
\newblock PhD thesis, 2019.

\bibitem{xu2017video}
Dejing Xu, Zhou Zhao, Jun Xiao, Fei Wu, Hanwang Zhang, Xiangnan He, and Yueting Zhuang.
\newblock Video question answering via gradually refined attention over appearance and motion.
\newblock In {\em ACM Multimedia}, 2017.

\bibitem{NEURIPS2019_c74d97b0}
Jiasen Lu, Dhruv Batra, Devi Parikh, and Stefan Lee.
\newblock Vilbert: Pretraining task-agnostic visiolinguistic representations for vision-and-language tasks.
\newblock In H.~Wallach, H.~Larochelle, A.~Beygelzimer, F.~d\textquotesingle Alch\'{e}-Buc, E.~Fox, and R.~Garnett, editors, {\em Advances in Neural Information Processing Systems}, volume~32. Curran Associates, Inc., 2019.

\bibitem{8839734}
Ting Yu, Jun Yu, Zhou Yu, and Dacheng Tao.
\newblock Compositional attention networks with two-stream fusion for video question answering.
\newblock {\em IEEE Transactions on Image Processing}, 29:1204--1218, 2020.

\bibitem{li2019visualbertsimpleperformantbaseline}
Liunian~Harold Li, Mark Yatskar, Da~Yin, Cho-Jui Hsieh, and Kai-Wei Chang.
\newblock Visualbert: A simple and performant baseline for vision and language, 2019.

\bibitem{7298698}
Fabian~Caba Heilbron, Victor Escorcia, Bernard Ghanem, and Juan~Carlos Niebles.
\newblock Activitynet: A large-scale video benchmark for human activity understanding.
\newblock In {\em 2015 IEEE Conference on Computer Vision and Pattern Recognition (CVPR)}, pages 961--970, 2015.

\bibitem{li2016tgifnewdatasetbenchmark}
Yuncheng Li, Yale Song, Liangliang Cao, Joel Tetreault, Larry Goldberg, Alejandro Jaimes, and Jiebo Luo.
\newblock Tgif: A new dataset and benchmark on animated gif description, 2016.

\bibitem{xiao2021nextqanextphasequestionansweringexplaining}
Junbin Xiao, Xindi Shang, Angela Yao, and Tat-Seng Chua.
\newblock Next-qa:next phase of question-answering to explaining temporal actions, 2021.

\bibitem{geminiteam2024geminifamilyhighlycapable}
Gemini~Team et~al.
\newblock Gemini: A family of highly capable multimodal models, 2024.

\bibitem{zhang2024videoinstructiontuningsynthetic}
Yuanhan Zhang, Jinming Wu, Wei Li, Bo~Li, Zejun Ma, Ziwei Liu, and Chunyuan Li.
\newblock Video instruction tuning with synthetic data, 2024.

\bibitem{grauman2022ego4dworld3000hours}
Kristen~Grauman et~al.
\newblock Ego4d: Around the world in 3,000 hours of egocentric video, 2022.

\bibitem{egoschema}
Karttikeya Mangalam, Raiymbek Akshulakov, and Jitendra Malik.
\newblock Egoschema: A diagnostic benchmark for very long-form video language understanding.
\newblock In A.~Oh, T.~Naumann, A.~Globerson, K.~Saenko, M.~Hardt, and S.~Levine, editors, {\em Advances in Neural Information Processing Systems}, volume~36, pages 46212--46244. Curran Associates, Inc., 2023.

\bibitem{rawal2024cinepile}
Ruchit Rawal, Khalid Saifullah, Ronen Basri, David Jacobs, Gowthami Somepalli, and Tom Goldstein.
\newblock Cinepile: A long video question answering dataset and benchmark.
\newblock {\em arXiv preprint arXiv:2405.08813}, 2024.

\bibitem{Rahman_2024_WACV}
Md~Motiur Rahman, Shiva Shokouhmand, Smriti Bhatt, and Miad Faezipour.
\newblock Mist: Medical image segmentation transformer with convolutional attention mixing (cam) decoder.
\newblock In {\em Proceedings of the IEEE/CVF Winter Conference on Applications of Computer Vision (WACV)}, pages 404--413, January 2024.

\bibitem{NIPS2014_d516b136}
Mateusz Malinowski and Mario Fritz.
\newblock A multi-world approach to question answering about real-world scenes based on uncertain input.
\newblock In Z.~Ghahramani, M.~Welling, C.~Cortes, N.~Lawrence, and K.Q. Weinberger, editors, {\em Advances in Neural Information Processing Systems}, volume~27. Curran Associates, Inc., 2014.

\bibitem{awad2023trecvid}
G.~Awad, K.~Curtis, A.~A. Butt, J.~Fiscus, A.~Godil, Y.~Lee, A.~Delgado, E.~Godard, L.~Diduch, D.~Gupta, D.~D. Fushman, Y.~Graham, and G.~Qu'enot.
\newblock Trecvid 2023 - a series of evaluation tracks in video understanding.
\newblock In {\em Proceedings of TRECVID 2023}. NIST, USA, 2023.

\bibitem{Xu_2021_CVPR}
Li~Xu, He~Huang, and Jun Liu.
\newblock Sutd-trafficqa: A question answering benchmark and an efficient network for video reasoning over traffic events.
\newblock In {\em Proceedings of the IEEE/CVF Conference on Computer Vision and Pattern Recognition (CVPR)}, pages 9878--9888, June 2021.

\bibitem{argosvision}
Argos Vision.
\newblock Argos vision: Advanced computer vision solutions, 2024.
\newblock Accessed: 2024-11-26.

\bibitem{xiao2024can}
Junbin Xiao, Angela Yao, Yicong Li, and Tat-Seng Chua.
\newblock Can i trust your answer? visually grounded video question answering.
\newblock In {\em Proceedings of the IEEE/CVF Conference on Computer Vision and Pattern Recognition}, pages 13204--13214, 2024.

\bibitem{li2024llava}
Bo~Li, Yuanhan Zhang, Dong Guo, Renrui Zhang, Feng Li, Hao Zhang, Kaichen Zhang, Yanwei Li, Ziwei Liu, and Chunyuan Li.
\newblock Llava-onevision: Easy visual task transfer.
\newblock {\em arXiv preprint arXiv:2408.03326}, 2024.

\bibitem{li2024llavanext}
Feng Li, Renrui Zhang, Hao Zhang, Yuanhan Zhang, Bo~Li, Wei Li, Zejun Ma, and Chunyuan Li.
\newblock Llava-next-interleave: Tackling multi-image, video, and 3d in large multimodal models.
\newblock {\em arXiv preprint arXiv:2407.07895}, 2024.

\bibitem{chen2024far}
Zhe Chen, Weiyun Wang, Hao Tian, Shenglong Ye, Zhangwei Gao, Erfei Cui, Wenwen Tong, Kongzhi Hu, Jiapeng Luo, Zheng Ma, et~al.
\newblock How far are we to gpt-4v? closing the gap to commercial multimodal models with open-source suites.
\newblock {\em arXiv preprint arXiv:2404.16821}, 2024.

\bibitem{cheng2024videollama2advancingspatialtemporal}
Zesen Cheng, Sicong Leng, Hang Zhang, Yifei Xin, Xin Li, Guanzheng Chen, Yongxin Zhu, Wenqi Zhang, Ziyang Luo, Deli Zhao, and Lidong Bing.
\newblock Videollama 2: Advancing spatial-temporal modeling and audio understanding in video-llms, 2024.

\bibitem{dong2024internlm}
Xiaoyi~Dong et~al.
\newblock Internlm-xcomposer2: Mastering free-form text-image composition and comprehension in vision-language large model.
\newblock {\em arXiv preprint arXiv:2401.16420}, 2024.

\bibitem{liu2023visualinstructiontuning}
Haotian~Liu et~al.
\newblock Improved baselines with visual instruction tuning.
\newblock In {\em NeurIPS}, volume~36, 2023.

\bibitem{lu2021interpretablegeometry}
Pan~Lu et~al.
\newblock Inter-gps: Interpretable geometry problem solving with formal language and symbolic reasoning.
\newblock {\em arXiv preprint arXiv:2105.04165}, 2021.

\bibitem{chen2023internvl}
Zhe Chen, Jiannan Wu, Wenhai Wang, Weijie Su, Guo Chen, Sen Xing, Muyan Zhong, Qinglong Zhang, Xizhou Zhu, Lewei Lu, Bin Li, Ping Luo, Tong Lu, Yu~Qiao, and Jifeng Dai.
\newblock Internvl: Scaling up vision foundation models and aligning for generic visual-linguistic tasks.
\newblock {\em arXiv preprint arXiv:2312.14238}, 2023.

\bibitem{cai2024internlm2}
Zheng~Cai et~al.
\newblock Internlm2 technical report.
\newblock {\em arXiv preprint arXiv:2403.17297}, 2024.

\bibitem{damonlpsg2023videollama}
Hang Zhang, Xin Li, and Lidong Bing.
\newblock Video-llama: An instruction-tuned audio-visual language model for video understanding.
\newblock {\em arXiv preprint arXiv:2306.02858}, 2023.

\bibitem{mathew2021docvqa}
et~al. Minesh~Mathew.
\newblock Docvqa: A dataset for vqa on document images.
\newblock In {\em WACV}, pages 2200--2209, 2021.

\bibitem{masry2022chartqa}
Ahmed~Masry et. al.
\newblock Chartqa: A benchmark for question answering about charts with visual and logical reasoning.
\newblock In {\em ACL}, pages 2263--2279, 2022.

\bibitem{mathew2022infographic}
Minesh~Mathew et~al.
\newblock Infographicvqa.
\newblock In {\em WACV}, pages 1697--1706, 2022.

\bibitem{singh2019towards}
Amanpreet~Singh et~al.
\newblock Towards vqa models that can read.
\newblock In {\em CVPR}, pages 8317--8326, 2019.

\bibitem{liu2023ocrbench}
Yuliang~Liu et~al.
\newblock On the hidden mystery of ocr in large multimodal models.
\newblock {\em arXiv preprint arXiv:2305.07895}, 2023.

\bibitem{mmtbench}
Kaining Ying, Fanqing Meng, Jin Wang, Zhiqian Li, Han Lin, Yue Yang, Hao Zhang, Wenbo Zhang, Yuqi Lin, Shuo Liu, Jiayi Lei, Quanfeng Lu, Runjian Chen, Peng Xu, Renrui Zhang, Haozhe Zhang, Peng Gao, Yali Wang, Yu~Qiao, Ping Luo, Kaipeng Zhang, and Wenqi Shao.
\newblock Mmt-bench: A comprehensive multimodal benchmark for evaluating large vision-language models towards multitask agi, 2024.

\bibitem{radford2021learning}
Alec Radford, Jong~Wook Kim, Chris Hallacy, Aditya Ramesh, Gabriel Goh, Sandhini Agarwal, Girish Sastry, Amanda Askell, Pamela Mishkin, Jack Clark, et~al.
\newblock Learning transferable visual models from natural language supervision.
\newblock In {\em International conference on machine learning}, pages 8748--8763. PMLR, 2021.

\bibitem{chen2023beats}
Chen et~al.
\newblock Beats: Audio pre-training with acoustic tokenizers.
\newblock In Andreas Krause, Emma Brunskill, Kyunghyun Cho, Barbara Engelhardt, Sivan Sabato, and Jonathan Scarlett, editors, {\em Proceedings of the 40th International Conference on Machine Learning}, volume 202 of {\em Proceedings of Machine Learning Research}, pages 5178--5193, 23--29 Jul 2023, July 2023. PMLR.

\bibitem{jiang2023mistral7b}
Albert~Q. Jiang, Alexandre Sablayrolles, Arthur Mensch, Chris Bamford, Devendra~Singh Chaplot, Diego de~las Casas, Florian Bressand, Gianna Lengyel, Guillaume Lample, Lucile Saulnier, Lélio~Renard Lavaud, Marie-Anne Lachaux, Pierre Stock, Teven~Le Scao, Thibaut Lavril, Thomas Wang, Timothée Lacroix, and William~El Sayed.
\newblock Mistral 7b, 2023.

\bibitem{chen2024panda70m}
Chen et~al.
\newblock Panda-70m: Captioning 70m videos with multiple cross-modality teachers.
\newblock {\em arXiv preprint arXiv:2402.19479}, 2024.

\bibitem{zhu2023languagebind}
Zhu et~al.
\newblock Languagebind: Extending video-language pretraining to n-modality by language-based semantic alignment.
\newblock {\em arXiv preprint arXiv:2310.01852}, 2023.

\bibitem{bain2021frozen}
Bain et~al.
\newblock Frozen in time: A joint video and image encoder for end-to-end retrieval.
\newblock In {\em Proceedings of the IEEE/CVF International Conference on Computer Vision (ICCV)}, 2021.

\bibitem{wang2023internvid}
Wang et~al.
\newblock Internvid: A large-scale video-text dataset for multimodal understanding and generation.
\newblock {\em arXiv preprint arXiv:2307.06942}, 2023.

\bibitem{changpinyo2021conceptual}
Changpinyo et~al.
\newblock Conceptual 12m: Pushing web-scale image-text pre-training to recognize long-tail visual concepts.
\newblock In {\em Proceedings of the IEEE/CVF Conference on Computer Vision and Pattern Recognition (CVPR)}, pages 3558--3568, 2021.

\bibitem{urbanek2023picture}
Urbanek et~al.
\newblock A picture is worth more than 77 text tokens: Evaluating clip-style models on dense captions.
\newblock {\em arXiv preprint arXiv:2312.08578}, 2023.

\bibitem{xue2022hdvila}
Hongwei Xue, Tiankai Hang, Yanhong Zeng, Yuchong Sun, Bei Liu, Huan Yang, Jianlong Fu, and Baining Guo.
\newblock Advancing high-resolution video-language representation with large-scale video transcriptions.
\newblock In {\em International Conference on Computer Vision and Pattern Recognition (CVPR)}, 2022.

\bibitem{vicuna2023}
Wei-Lin Chiang, Zhuohan Li, Zi~Lin, Ying Sheng, Zhanghao Wu, Hao Zhang, Lianmin Zheng, Siyuan Zhuang, Yonghao Zhuang, Joseph~E. Gonzalez, Ion Stoica, and Eric~P. Xing.
\newblock Vicuna: An open-source chatbot impressing gpt-4 with 90\%* chatgpt quality, March 2023.

\bibitem{liu2024llavanext}
Haotian Liu, Chunyuan Li, Yuheng Li, Bo~Li, Yuanhan Zhang, Sheng Shen, and Yong~Jae Lee.
\newblock Llava-next: Improved reasoning, ocr, and world knowledge, January 2024.

\bibitem{li2024llavanext-strong}
Bo~Li, Kaichen Zhang, Hao Zhang, Dong Guo, Renrui Zhang, Feng Li, Yuanhan Zhang, Ziwei Liu, and Chunyuan Li.
\newblock Llava-next: Stronger llms supercharge multimodal capabilities in the wild, May 2024.

\end{thebibliography}

\newpage
\appendix
\section{Appendix}
\label{appendix}
\subsection{Question Prompt Generation}

Below is a sample prompt provided to GPT for curating questions. The prompt begins with user-curated questions, followed by a description of the traffic scenarios, which are meant to be detailed and elaborate.
The prompt focuses on analyzing a traffic scenario captured in a video of a busy intersection. In the video, a cyclist is seen riding on the wrong side of the road, violating traffic regulations. Instead of stopping at the intersection as required, the cyclist speeds through, creating a potential traffic hazard. The video duration spans from $8s$ to $10s$, and questions are curated to evaluate observations and understanding at different levels of complexity.
The curated questions are designed to assess spatial awareness, logical reasoning, and traffic engineering insights, providing a comprehensive evaluation of the scenario.

\textbf{\textit{Task Description:}}
\textit{
    You are an expert in linguistics, logic, and traffic engineering, tasked with creating questions based on three given traffic scenarios. The questions will assess understanding at different difficulty levels, focusing on spatial and temporal aspects. The questions will be categorized as follows:
}

\begin{itemize}
    \item \textbf{\textit{Easy}}: Questions that focus on straightforward observations (e.g., "How many cars can you see?").
    \item \textbf{\textit{Moderate}}: Questions requiring attention to detail or basic reasoning (e.g., "Is the cyclist following the bike lane?").
    \item \textbf{\textit{Hard}}: Questions involving detailed analysis or traffic engineering knowledge (e.g., "What impact does the pedestrian's position have on vehicle flow?").
\end{itemize}

\textbf{\textit{Output Requirements:}}

\begin{itemize}
    \item \textit{The result should be in \textbf{JSON format}}.
    \item \textit{Each question must have a \textbf{unique ID}.}
    \item \textit{Generate \textbf{10 questions} for each category (\textbf{30 questions in total})}.
\end{itemize}
\textbf{\textit{Scenario Details:}}

\textit{Provide the scenario details here.}

\textbf{\textit{Busy Intersection}}

\textit{
\textbf{Situation:} A cyclist is on the wrong side of the road. In the video, the cyclist zips through the intersection without stopping, creating a potential traffic hazard.
}

\textbf{\textit{Questions Curated:}}

\begin{itemize}
    \item \textbf{\textit{Easy}}:
    \begin{itemize}
        \item Is the video taken in daylight or at night?
        \item How many cyclists are there?
        \item How many cars are there?
    \end{itemize}
    
    \item \textbf{\textit{Moderate}}:
    \begin{itemize}
        \item Is the cyclist following the bike lane?
    \end{itemize}
    
    \item \textbf{\textit{Hard}}:
    \begin{itemize}
        \item Are there any visible landmarks?
        \item Between 0:08 to 0:10, how many cars are moving?
    \end{itemize}
\end{itemize}
\textbf{\textit{Question Requirements:}}

\begin{itemize}
    \item \textbf{\textit{Easy}}: Questions should be simple and basic, focusing on straightforward observations.
    \item \textbf{\textit{Moderate}}: Questions should require some attention to detail or involve interpreting specific behaviors or interactions.
    \item \textbf{\textit{Hard}}: Questions should involve advanced reasoning, detailed analysis, or traffic engineering knowledge, with timestamps implied but not explicitly stated.
\end{itemize}

\textit{
\textbf{Note:} Do not generate the answers.
}
   
\subsection{Question Sets Used to Evaluate VideoQA Models}
As discussed earlier (Section \ref{method-of-study} and \ref{experiment}), the question sets (Table \ref{allQuestionsSeq1},  \ref{allQuestionsSeq2}, and \ref{allQuestionsSeq3}) used to evaluate the SOTA VideoQA models used in this study and their performance  were analyzed for each type of question. All questions are presented with incorrect answers highlighted in \textcolor{red}{red} and correct ones in \textcolor{green}{green}, comparing against the ground truth. The evaluation steps and the code used during this study is made available at \url{https://github.com/joe-rabbit/VideoQA_Pilot_Study}.

\begin{table}[h]
\centering
\caption{Question sets used on real-world traffic video sequence 1}
\label{allQuestionsSeq1}
\resizebox{\columnwidth}{!}{%
\begin{tabular}{
>{\columncolor[HTML]{FFFFFF}}c |
>{\columncolor[HTML]{FFFFFF}}l |
>{\columncolor[HTML]{9AFF99}}l 
>{\columncolor[HTML]{FFCCC9}}l l
>{\columncolor[HTML]{FFCCC9}}l 
>{\columncolor[HTML]{9AFF99}}l |l}
\cline{2-7}
\multicolumn{1}{l|}{\cellcolor[HTML]{FFFFFF}}
&
  \multicolumn{6}{c|}{\cellcolor[HTML]{FFFFFF}\includegraphics[width=1.5\textwidth]{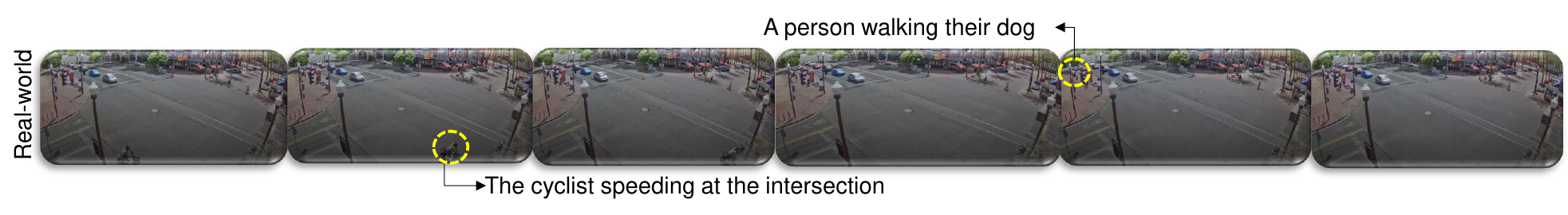}}  \\  \cline{2-7}

&
  \multicolumn{1}{c|}{\cellcolor[HTML]{EFEFEF}} &
  \multicolumn{5}{c|}{\cellcolor[HTML]{EFEFEF}\textbf{Video QA model Answers}} &
   \\ \cline{3-7}
\multicolumn{1}{l|}{\multirow{-2}{*}{\cellcolor[HTML]{FFFFFF}}} &
  \multicolumn{1}{c|}{\multirow{-2}{*}{\cellcolor[HTML]{EFEFEF}\textbf{Questions}}} &
  \multicolumn{1}{c|}{\cellcolor[HTML]{EFEFEF}\textbf{ChatGPT}} &
  \multicolumn{1}{c|}{\cellcolor[HTML]{EFEFEF}\textbf{LLaVa}} &
  \multicolumn{1}{c|}{\cellcolor[HTML]{EFEFEF}\textbf{Gemini}} &
  \multicolumn{1}{c|}{\cellcolor[HTML]{EFEFEF}\textbf{InternVL}} &
  \multicolumn{1}{c|}{\cellcolor[HTML]{EFEFEF}\textbf{VideoLLaMA-2}} &
   \\ \cmidrule{1-7} 
   \cline{1-7}
\multicolumn{1}{|c|}{\cellcolor[HTML]{EFEFEF}} &
  \begin{tabular}[c]{@{}l@{}}Is the cyclist following the\\ bike lane?\end{tabular} &
  \multicolumn{1}{l|}{\cellcolor[HTML]{9AFF99}\begin{tabular}[c]{@{}l@{}}The cyclist is not following\\ the bike lane properly.\end{tabular}} &
  \multicolumn{1}{l|}{\cellcolor[HTML]{FFCCC9}\begin{tabular}[c]{@{}l@{}}The cyclist appears to\\ be following the bike lane.\end{tabular}} &
  \multicolumn{1}{l|}{\cellcolor[HTML]{FFCCC9}\begin{tabular}[c]{@{}l@{}}No, the cyclist is riding\\ against the flow of traffic\\ in the bike lane.\end{tabular}} &
  \multicolumn{1}{l|}{\cellcolor[HTML]{FFCCC9}\begin{tabular}[c]{@{}l@{}}Yes, the cyclist is in the\\ bike lane.\end{tabular}} &
  \begin{tabular}[c]{@{}l@{}}The cyclist is not following\\ the bike lane.\end{tabular} &
   \\ \cline{2-7}
\multicolumn{1}{|c|}{\cellcolor[HTML]{EFEFEF}} &
  \begin{tabular}[c]{@{}l@{}}Are there any pedestrians in\\ the video?\end{tabular} &
  \multicolumn{1}{l|}{\cellcolor[HTML]{FFCCC9}\begin{tabular}[c]{@{}l@{}}No, there are no visible\\ pedestrians in the video.\end{tabular}} &
  \multicolumn{1}{l|}{\cellcolor[HTML]{9AFF99}\begin{tabular}[c]{@{}l@{}}Yes, there are pedestrians\\ on the sidewalk and crossing.\end{tabular}} &
  \multicolumn{1}{l|}{\cellcolor[HTML]{9AFF99}\begin{tabular}[c]{@{}l@{}}Yes, two pedestrians are\\ visible.\end{tabular}} &
  \multicolumn{1}{l|}{\cellcolor[HTML]{9AFF99}\begin{tabular}[c]{@{}l@{}}A woman walks on the\\ sidewalk and a person in\\ the street.\end{tabular}} &
  \begin{tabular}[c]{@{}l@{}}Yes, pedestrians are visible\\ in some frames.\end{tabular} &
   \\ \cline{2-7}
\multicolumn{1}{|c|}{\cellcolor[HTML]{EFEFEF}} &
  \begin{tabular}[c]{@{}l@{}}Is there a traffic light in the\\ video?\end{tabular} &
  \multicolumn{1}{l|}{\cellcolor[HTML]{9AFF99}\begin{tabular}[c]{@{}l@{}}Yes, there is a traffic\\ light visible.\end{tabular}} &
  \multicolumn{1}{l|}{\cellcolor[HTML]{9AFF99}\begin{tabular}[c]{@{}l@{}}Yes, a red traffic light is\\ visible.\end{tabular}} &
  \multicolumn{1}{l|}{\cellcolor[HTML]{9AFF99}\begin{tabular}[c]{@{}l@{}}Yes, on the left side of\\ the frame.\end{tabular}} &
  \multicolumn{1}{l|}{\cellcolor[HTML]{9AFF99}\begin{tabular}[c]{@{}l@{}}Yes, located on a traffic\\ pole at the intersection.\end{tabular}} &
  \begin{tabular}[c]{@{}l@{}}Yes, a red traffic light is\\ visible when the car is stopped.\end{tabular} &
   \\ \cline{2-7}
\multicolumn{1}{|c|}{\cellcolor[HTML]{EFEFEF}} &
  \begin{tabular}[c]{@{}l@{}}Is there a person walking a\\ dog?\end{tabular} &
  \multicolumn{1}{l|}{\cellcolor[HTML]{FFCCC9}\begin{tabular}[c]{@{}l@{}}No, there is no person\\ walking a dog.\end{tabular}} &
  \multicolumn{1}{l|}{\cellcolor[HTML]{9AFF99}\begin{tabular}[c]{@{}l@{}}Yes, a person is walking\\ a dog on the sidewalk.\end{tabular}} &
  \multicolumn{1}{l|}{\cellcolor[HTML]{FFCCC9}No.} &
  \multicolumn{1}{l|}{\cellcolor[HTML]{9AFF99}\begin{tabular}[c]{@{}l@{}}Yes, a person with a dog\\ is visible near the corner.\end{tabular}} &
  \begin{tabular}[c]{@{}l@{}}Yes, there is a person\\ walking a dog in the video.\end{tabular} &
   \\ \cline{2-7}
\multicolumn{1}{|c|}{\cellcolor[HTML]{EFEFEF}} &
  \begin{tabular}[c]{@{}l@{}}How many cyclists are\\ there?\end{tabular} &
  \multicolumn{1}{l|}{\cellcolor[HTML]{9AFF99}There is one cyclist.} &
  \multicolumn{1}{l|}{\cellcolor[HTML]{FFCCC9}\begin{tabular}[c]{@{}l@{}}No cyclists visible in the\\ image.\end{tabular}} &
  \multicolumn{1}{l|}{\cellcolor[HTML]{9AFF99}One.} &
  \multicolumn{1}{l|}{\cellcolor[HTML]{9AFF99}\begin{tabular}[c]{@{}l@{}}One cyclist is visible in\\ the video.\end{tabular}} &
  \cellcolor[HTML]{FFCCC9}\begin{tabular}[c]{@{}l@{}}There a couple of cyclists in \\ the video\end{tabular} &
   \\ \cline{2-7}
\multicolumn{1}{|c|}{\multirow{-10}{*}{\rotatebox[origin=c]{90}{\centering {\cellcolor[HTML]{EFEFEF}\textbf{Easy}}}}} &
  \begin{tabular}[c]{@{}l@{}}Is there no person walking a\\ dog?\end{tabular} &
  \multicolumn{1}{l|}{\cellcolor[HTML]{FFCCC9}\begin{tabular}[c]{@{}l@{}}No, there is no person\\ walking a dog in the\\ video.\end{tabular}} &
  \multicolumn{1}{l|}{\cellcolor[HTML]{FFCCC9}\begin{tabular}[c]{@{}l@{}}Yes, there is no person\\ walking a dog.\end{tabular}} &
  \multicolumn{1}{l|}{\cellcolor[HTML]{FFCCC9}\begin{tabular}[c]{@{}l@{}}No, there is no person\\ walking a dog.\end{tabular}} &
  \multicolumn{1}{l|}{\cellcolor[HTML]{FFCCC9}\begin{tabular}[c]{@{}l@{}}No person walking a\\ dog in any part of the\\ footage.\end{tabular}} &
  \begin{tabular}[c]{@{}l@{}}No, there is a person\\ walking a dog.\end{tabular} &
   \\ \cmidrule{1-7} 
   \cline{1-7}
\multicolumn{1}{|c|}{\cellcolor[HTML]{EFEFEF}} &
  \begin{tabular}[c]{@{}l@{}}Was there any vehicle that\\ signaled before turning?\end{tabular} &
  \multicolumn{1}{l|}{\cellcolor[HTML]{9AFF99}\begin{tabular}[c]{@{}l@{}}No, there were no visible\\ signals before turning.\end{tabular}} &
  \multicolumn{1}{l|}{\cellcolor[HTML]{FFCCC9}\begin{tabular}[c]{@{}l@{}}Yes, a car signaled before\\ turning.\end{tabular}} &
  \multicolumn{1}{l|}{\cellcolor[HTML]{9AFF99}No.} &
  \multicolumn{1}{l|}{\cellcolor[HTML]{FFCCC9}\begin{tabular}[c]{@{}l@{}}Yes, a silver car signaled\\ before turning.\end{tabular}} &
  \cellcolor[HTML]{FFCCC9}\begin{tabular}[c]{@{}l@{}}Yes, a white car signaled\\ before turning.\end{tabular} &
   \\ \cline{2-7}
\multicolumn{1}{|c|}{\cellcolor[HTML]{EFEFEF}} &
  \begin{tabular}[c]{@{}l@{}}How many cars were in\\ motion while the cyclist\\ crossed?\end{tabular} &
  \multicolumn{1}{l|}{\cellcolor[HTML]{9AFF99}Two cars were in motion.} &
  \multicolumn{1}{l|}{\cellcolor[HTML]{9AFF99}Two cars were moving.} &
  \multicolumn{1}{l|}{\cellcolor[HTML]{FFCCC9}Three cars were in motion.} &
  \multicolumn{1}{l|}{\cellcolor[HTML]{9AFF99}\begin{tabular}[c]{@{}l@{}}Two cars were moving\\ as the cyclist crossed.\end{tabular}} &
  \begin{tabular}[c]{@{}l@{}}Two cars were in motion\\ while the cyclist crossed.\end{tabular} &
   \\ \cline{2-7}
\multicolumn{1}{|c|}{\cellcolor[HTML]{EFEFEF}} &
  \begin{tabular}[c]{@{}l@{}}How many vehicles are\\ stopped at the intersection\\ when the cyclist crosses?\end{tabular} &
  \multicolumn{1}{l|}{\cellcolor[HTML]{FFCCC9}\begin{tabular}[c]{@{}l@{}}Three vehicles are\\ stopped at the intersection.\end{tabular}} &
  \multicolumn{1}{l|}{\cellcolor[HTML]{FFCCC9}Two vehicles are stopped.} &
  \multicolumn{1}{l|}{\cellcolor[HTML]{FFCCC9}No vehicles stopped.} &
  \multicolumn{1}{l|}{\cellcolor[HTML]{FFCCC9}\begin{tabular}[c]{@{}l@{}}Three vehicles are\\ stopped.\end{tabular}} &
  \cellcolor[HTML]{FFCCC9}\begin{tabular}[c]{@{}l@{}}Two vehicles are\\ stopped.\end{tabular} &
   \\ \cline{2-7}
\multicolumn{1}{|c|}{\cellcolor[HTML]{EFEFEF}} &
  \begin{tabular}[c]{@{}l@{}}Which vehicle is closest to\\ the cyclist when they pass\\ through the intersection?\end{tabular} &
  \multicolumn{1}{l|}{\cellcolor[HTML]{9AFF99}A sedan is the closest vehicle.} &
  \multicolumn{1}{l|}{\cellcolor[HTML]{FFCCC9}\begin{tabular}[c]{@{}l@{}}A car on the right side is\\ closest to the cyclist.\end{tabular}} &
  \multicolumn{1}{l|}{\cellcolor[HTML]{FFCCC9}\begin{tabular}[c]{@{}l@{}}The red car is closest to\\ the cyclist.\end{tabular}} &
  \multicolumn{1}{l|}{\cellcolor[HTML]{9AFF99}\begin{tabular}[c]{@{}l@{}}The white car is closest\\ to the cyclist.\end{tabular}} &
  The white car is closest. &
   \\ \cline{2-7}
\multicolumn{1}{|c|}{\cellcolor[HTML]{EFEFEF}} &
  \begin{tabular}[c]{@{}l@{}}Was there a pedestrian \\ near the cyclist when they\\ crossed?\end{tabular} &
  \multicolumn{1}{l|}{\cellcolor[HTML]{9AFF99}\begin{tabular}[c]{@{}l@{}}No, there was no pedestrian\\ near the cyclist.\end{tabular}} &
  \multicolumn{1}{l|}{\cellcolor[HTML]{9AFF99}\begin{tabular}[c]{@{}l@{}}No pedestrian near the\\ cyclist.\end{tabular}} &
  \multicolumn{1}{l|}{\cellcolor[HTML]{FFCCC9}\begin{tabular}[c]{@{}l@{}}Yes, there is a pedestrian\\ on the sidewalk.\end{tabular}} &
  \multicolumn{1}{l|}{\cellcolor[HTML]{FFCCC9}\begin{tabular}[c]{@{}l@{}}One pedestrian was near\\ the cyclist.\end{tabular}} &
  \cellcolor[HTML]{FFCCC9}\begin{tabular}[c]{@{}l@{}}Yes, there was a pedestrian\\ near the cyclist.\end{tabular} &
   \\ \cline{2-7}
\multicolumn{1}{|c|}{\multirow{-10}{*}{\rotatebox[origin=c]{90}{\centering {\cellcolor[HTML]{EFEFEF}\textbf{Moderate}}}}} &
  \begin{tabular}[c]{@{}l@{}}Was there no pedestrian\\ near the cyclist when they\\ crossed?\end{tabular} &
  \multicolumn{1}{l|}{\cellcolor[HTML]{9AFF99}\begin{tabular}[c]{@{}l@{}}Yes, there was no pedestrian\\ near the cyclist.\end{tabular}} &
  \multicolumn{1}{l|}{\cellcolor[HTML]{9AFF99}\begin{tabular}[c]{@{}l@{}}Yes, there was no pedestrian\\ near the cyclist.\end{tabular}} &
  \multicolumn{1}{l|}{\cellcolor[HTML]{9AFF99}\begin{tabular}[c]{@{}l@{}}No pedestrian was near\\ the cyclist.\end{tabular}} &
  \multicolumn{1}{l|}{\cellcolor[HTML]{FFCCC9}\begin{tabular}[c]{@{}l@{}}There were people on\\ the sidewalk.\end{tabular}} &
  \begin{tabular}[c]{@{}l@{}}Yes, there was no pedestrian\\ near the cyclist.\end{tabular} &
   \\ \cmidrule{1-7} 
   \cline{1-7}
\multicolumn{1}{|c|}{\cellcolor[HTML]{EFEFEF}} &
  \begin{tabular}[c]{@{}l@{}}Between 0:08 and 0:10, how\\ many cars are moving?\end{tabular} &
  \multicolumn{1}{l|}{\cellcolor[HTML]{FFCCC9}Two cars are moving.} &
  \multicolumn{1}{l|}{\cellcolor[HTML]{FFCCC9}\begin{tabular}[c]{@{}l@{}}Two cars are moving in\\ the image.\end{tabular}} &
  \multicolumn{1}{l|}{\cellcolor[HTML]{FFCCC9}Four cars are moving.} &
  \multicolumn{1}{l|}{\cellcolor[HTML]{9AFF99}Only one car is moving.} &
  One car is moving. &
   \\ \cline{2-7}
\multicolumn{1}{|c|}{\cellcolor[HTML]{EFEFEF}} &
  \begin{tabular}[c]{@{}l@{}}How does the cyclist’s violation\\ impact overall traffic\\ safety?\end{tabular} &
  \multicolumn{1}{l|}{\cellcolor[HTML]{FFCCC9}\begin{tabular}[c]{@{}l@{}}Creates a potential hazard,\\ disrupting flow.\end{tabular}} &
  \multicolumn{1}{l|}{\cellcolor[HTML]{FFCCC9}\begin{tabular}[c]{@{}l@{}}Violating traffic rules\\ creates confusion and\\ congestion.\end{tabular}} &
  \multicolumn{1}{l|}{\cellcolor[HTML]{FFCCC9}Increases risk of collision.} &
  \multicolumn{1}{l|}{\cellcolor[HTML]{FFCCC9}\begin{tabular}[c]{@{}l@{}}Immediate disregard for\\ traffic regulations increases\\ risk.\end{tabular}} &
  \cellcolor[HTML]{FFCCC9}\begin{tabular}[c]{@{}l@{}}Running red light disrupts\\ flow and increases\\ risk.\end{tabular} &
   \\ \cline{2-7}
\multicolumn{1}{|c|}{\cellcolor[HTML]{EFEFEF}} &
  \begin{tabular}[c]{@{}l@{}}What design elements could\\ improve bicycle-vehicle segregation?\end{tabular} &
  \multicolumn{1}{l|}{\cellcolor[HTML]{9AFF99}\begin{tabular}[c]{@{}l@{}}Clearer bike lane markings\\ and barriers.\end{tabular}} &
  \multicolumn{1}{l|}{\cellcolor[HTML]{9AFF99}\begin{tabular}[c]{@{}l@{}}Separate lanes, signage,\\ and traffic calming measures.\end{tabular}} &
  \multicolumn{1}{l|}{\cellcolor[HTML]{9AFF99}\begin{tabular}[c]{@{}l@{}}Physically separated\\ bike lane.\end{tabular}} &
  \multicolumn{1}{l|}{\cellcolor[HTML]{FFCCC9}\begin{tabular}[c]{@{}l@{}}Barriers or delineators\\ could improve separation.\end{tabular}} &
  \begin{tabular}[c]{@{}l@{}}Dedicated bike lanes,\\ bike boxes, and barriers.\end{tabular} &
   \\ \cline{2-7}
\multicolumn{1}{|c|}{\cellcolor[HTML]{EFEFEF}} &
  \begin{tabular}[c]{@{}l@{}}What impact does the cyclist’s\\ position have on vehicle\\ flow?\end{tabular} &
  \multicolumn{1}{l|}{\cellcolor[HTML]{FFCCC9}\begin{tabular}[c]{@{}l@{}}Forces vehicles to slow\\ down, increasing risk.\end{tabular}} &
  \multicolumn{1}{l|}{\cellcolor[HTML]{FFCCC9}\begin{tabular}[c]{@{}l@{}}Cyclist in intersection\\ may cause delays.\end{tabular}} &
  \multicolumn{1}{l|}{\cellcolor[HTML]{9AFF99}\begin{tabular}[c]{@{}l@{}}Minimal impact in current\\ instance.\end{tabular}} &
  \multicolumn{1}{l|}{\cellcolor[HTML]{FFCCC9}\begin{tabular}[c]{@{}l@{}}Position reduces one\\ lane, affecting flow.\end{tabular}} &
  \cellcolor[HTML]{FFCCC9}\begin{tabular}[c]{@{}l@{}}Cyclist obstructs traffic\\ flow, causing congestion.\end{tabular} &
   \\ \cline{2-7}
\multicolumn{1}{|c|}{\cellcolor[HTML]{EFEFEF}} &
  \begin{tabular}[c]{@{}l@{}}Did the cyclist appear to\\ slow down when approaching?\end{tabular} &
  \multicolumn{1}{l|}{\cellcolor[HTML]{9AFF99}\begin{tabular}[c]{@{}l@{}}No, the cyclist did not\\ slow down.\end{tabular}} &
  \multicolumn{1}{l|}{\cellcolor[HTML]{FFCCC9}\begin{tabular}[c]{@{}l@{}}Yes, cyclist slows down\\ for safety.\end{tabular}} &
  \multicolumn{1}{l|}{\cellcolor[HTML]{9AFF99}No, steady pace.} &
  \multicolumn{1}{l|}{\cellcolor[HTML]{9AFF99}No indication of slowdown.} &
  \cellcolor[HTML]{FFCCC9}Yes, cyclist slows down. &
   \\ \cline{2-7}
\multicolumn{1}{|c|}{\multirow{-10}{*}{\rotatebox[origin=c]{90}{\centering {\cellcolor[HTML]{EFEFEF}\textbf{Complex}}}}} &
  \begin{tabular}[c]{@{}l@{}}Did the cyclist not appear to\\ slow down when approaching?\end{tabular} &
  \multicolumn{1}{l|}{\cellcolor[HTML]{9AFF99}\begin{tabular}[c]{@{}l@{}}Yes, cyclist did not slow\\ down.\end{tabular}} &
  \multicolumn{1}{l|}{\cellcolor[HTML]{9AFF99}\begin{tabular}[c]{@{}l@{}}Cyclist maintained\\ steady speed.\end{tabular}} &
  \multicolumn{1}{l|}{\cellcolor[HTML]{9AFF99}\begin{tabular}[c]{@{}l@{}}Cyclist did not slow\\ down.\end{tabular}} &
  \multicolumn{1}{l|}{\cellcolor[HTML]{FFCCC9}\begin{tabular}[c]{@{}l@{}}Video does not confirm\\ slowdown.\end{tabular}} &
  \begin{tabular}[c]{@{}l@{}}Yes, cyclist did not slow\\ down.\end{tabular} &
   \\ \cmidrule{1-7} 
   \cline{1-7}
\end{tabular}%
}
\end{table}

\begin{table}[]
\centering
\caption{Question sets used on real-world traffic video sequence 2}
\label{allQuestionsSeq2}
\resizebox{\columnwidth}{!}{%
\begin{tabular}{
>{\columncolor[HTML]{FFFFFF}}c |
>{\columncolor[HTML]{FFFFFF}}l |
>{\columncolor[HTML]{9AFF99}}l 
>{\columncolor[HTML]{FFCCC9}}l l
>{\columncolor[HTML]{FFCCC9}}l 
>{\columncolor[HTML]{9AFF99}}l |l}
\cline{2-7}
\multicolumn{1}{l|}{\cellcolor[HTML]{FFFFFF}}
&
  \multicolumn{6}{c|}{\cellcolor[HTML]{FFFFFF}\includegraphics[width=1.5\textwidth]{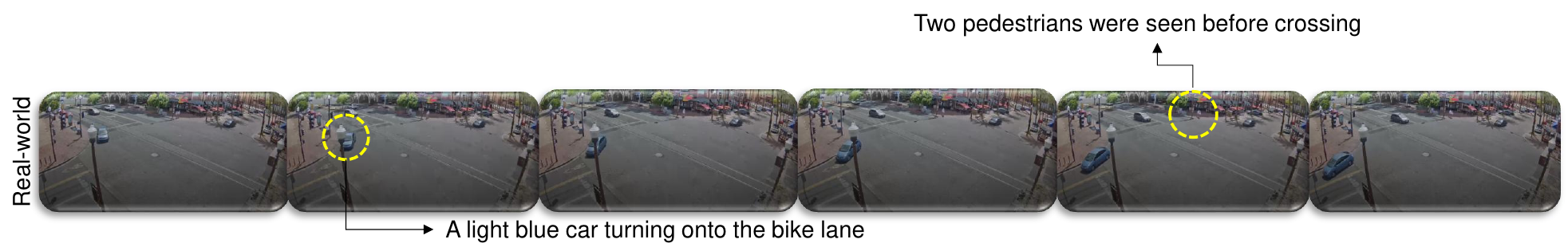}}  \\  \cline{2-7}

&
  \multicolumn{1}{c|}{\cellcolor[HTML]{EFEFEF}} &
  \multicolumn{5}{c|}{\cellcolor[HTML]{EFEFEF}\textbf{Video QA model Answers}} &
   \\ \cline{3-7}
\multicolumn{1}{l|}{\multirow{-2}{*}{\cellcolor[HTML]{FFFFFF}}} &
  \multicolumn{1}{c|}{\multirow{-2}{*}{\cellcolor[HTML]{EFEFEF}\textbf{Questions}}} &
  \multicolumn{1}{c|}{\cellcolor[HTML]{EFEFEF}\textbf{ChatGPT}} &
  \multicolumn{1}{c|}{\cellcolor[HTML]{EFEFEF}\textbf{LLaVa}} &
  \multicolumn{1}{c|}{\cellcolor[HTML]{EFEFEF}\textbf{Gemini}} &
  \multicolumn{1}{c|}{\cellcolor[HTML]{EFEFEF}\textbf{InternVL}} &
  \multicolumn{1}{c|}{\cellcolor[HTML]{EFEFEF}\textbf{VideoLLaMA-2}} 
   \\ \cmidrule{1-7} 
   \cline{1-7}
\multicolumn{1}{|c|}{\cellcolor[HTML]{EFEFEF}} & \begin{tabular}[c]{@{}l@{}}How many lamp posts are\\ visible?\end{tabular} & \multicolumn{1}{l|}{\cellcolor[HTML]{FFCCC9}Two lamp posts are visible.} & \multicolumn{1}{l|}{\cellcolor[HTML]{FFCCC9}\begin{tabular}[c]{@{}l@{}}There are two lamp\\ posts visible in the image.\end{tabular}} & \multicolumn{1}{l|}{\cellcolor[HTML]{9AFF99}\begin{tabular}[c]{@{}l@{}}Three lamp posts in\\ the foreground, several\\ more far back.\end{tabular}} & \multicolumn{1}{l|}{\cellcolor[HTML]{FFCCC9}\begin{tabular}[c]{@{}l@{}}Two lamp posts visible\\ on the left side.\end{tabular}} & \cellcolor[HTML]{FFCCC9}\begin{tabular}[c]{@{}l@{}}Four lamp posts in the\\ video.\end{tabular} &  \\ \cline{2-7}
\multicolumn{1}{|c|}{\cellcolor[HTML]{EFEFEF}} & \begin{tabular}[c]{@{}l@{}}What color is the first car\\ that turns onto the bike\\ lane?\end{tabular} & \multicolumn{1}{l|}{\cellcolor[HTML]{FFCCC9}The first car is white.} & \multicolumn{1}{l|}{\cellcolor[HTML]{FFCCC9}The first car is white.} & \multicolumn{1}{l|}{\cellcolor[HTML]{FFCCC9}White.} & \multicolumn{1}{l|}{\cellcolor[HTML]{FFCCC9}Black.} & \cellcolor[HTML]{FFCCC9}Red. &  \\ \cline{2-7}
\multicolumn{1}{|c|}{\cellcolor[HTML]{EFEFEF}} & \begin{tabular}[c]{@{}l@{}}How many pedestrians are\\ there?\end{tabular} & \multicolumn{1}{l|}{\cellcolor[HTML]{FFCCC9}No visible pedestrians.} & \multicolumn{1}{l|}{\cellcolor[HTML]{FFCCC9}Two pedestrians are visible.} & \multicolumn{1}{l|}{\cellcolor[HTML]{9AFF99}\begin{tabular}[c]{@{}l@{}}At least six different\\ pedestrians.\end{tabular}} & \multicolumn{1}{l|}{\cellcolor[HTML]{FFCCC9}\begin{tabular}[c]{@{}l@{}}Three at the beginning,\\ two toward the end.\end{tabular}} & \cellcolor[HTML]{FFCCC9}A couple of pedestrians. &  \\ \cline{2-7}
\multicolumn{1}{|c|}{\cellcolor[HTML]{EFEFEF}} & Are there any bicycles? & \multicolumn{1}{l|}{\cellcolor[HTML]{FFCCC9}No bicycles in the video.} & \multicolumn{1}{l|}{\cellcolor[HTML]{9AFF99}\begin{tabular}[c]{@{}l@{}}Yes, a bicycle parked on\\ the sidewalk.\end{tabular}} & \multicolumn{1}{l|}{\cellcolor[HTML]{9AFF99}\begin{tabular}[c]{@{}l@{}}Yes, a bicycle at the bottom\\ left.\end{tabular}} & \multicolumn{1}{l|}{\cellcolor[HTML]{FFCCC9}No bicycles visible.} & \cellcolor[HTML]{9AFF99}\begin{tabular}[c]{@{}l@{}}Yes, bicycles in the\\ video.\end{tabular} &  \\ \cline{2-7}
\multicolumn{1}{|c|}{\cellcolor[HTML]{EFEFEF}} & Are there any trucks? & \multicolumn{1}{l|}{\cellcolor[HTML]{FFCCC9}Yes, one truck visible.} & \multicolumn{1}{l|}{\cellcolor[HTML]{9AFF99}No trucks visible.} & \multicolumn{1}{l|}{\cellcolor[HTML]{9AFF99}No.} & \multicolumn{1}{l|}{\cellcolor[HTML]{9AFF99}No trucks present.} & \cellcolor[HTML]{FFCCC9}\begin{tabular}[c]{@{}l@{}}Yes, a white truck in the\\ video.\end{tabular} &  \\ \cline{2-7}
\multicolumn{1}{|c|}{\multirow{-10}{*}{\rotatebox[origin=c]{90}{\centering {\cellcolor[HTML]{EFEFEF}\textbf{Easy}}}}} & Are there no trucks? & \multicolumn{1}{l|}{\cellcolor[HTML]{9AFF99}\begin{tabular}[c]{@{}l@{}}Yes, there are no trucks\\ visible.\end{tabular}} & \multicolumn{1}{l|}{\cellcolor[HTML]{9AFF99}Yes, no trucks visible.} & \multicolumn{1}{l|}{\cellcolor[HTML]{9AFF99}No trucks.} & \multicolumn{1}{l|}{\cellcolor[HTML]{9AFF99}No trucks in the video.} & \cellcolor[HTML]{9AFF99}There are no trucks. &  \\ \cmidrule{1-7} 
   \cline{1-7}
\multicolumn{1}{|c|}{\cellcolor[HTML]{EFEFEF}} & \begin{tabular}[c]{@{}l@{}}How many cars turn onto\\ the bike lane?\end{tabular} & \multicolumn{1}{l|}{\cellcolor[HTML]{9AFF99}\begin{tabular}[c]{@{}l@{}}Three cars turn onto the\\ bike lane.\end{tabular}} & \multicolumn{1}{l|}{\cellcolor[HTML]{FFCCC9}\begin{tabular}[c]{@{}l@{}}Two cars turn onto the\\ bike lane.\end{tabular}} & \multicolumn{1}{l|}{\cellcolor[HTML]{FFCCC9}\begin{tabular}[c]{@{}l@{}}Six cars turn, two exit\\ quickly.\end{tabular}} & \multicolumn{1}{l|}{\cellcolor[HTML]{FFCCC9}\begin{tabular}[c]{@{}l@{}}Two cars turn into the\\ bike lane.\end{tabular}} & \cellcolor[HTML]{FFCCC9}One car turns. &  \\ \cline{2-7}
\multicolumn{1}{|c|}{\cellcolor[HTML]{EFEFEF}} & \begin{tabular}[c]{@{}l@{}}Is there a manhole cover\\ next to the tracks?\end{tabular} & \multicolumn{1}{l|}{\cellcolor[HTML]{FFCCC9}\begin{tabular}[c]{@{}l@{}}No, no manhole cover\\ visible near tracks.\end{tabular}} & \multicolumn{1}{l|}{\cellcolor[HTML]{9AFF99}\begin{tabular}[c]{@{}l@{}}Yes, manhole cover next\\ to tracks.\end{tabular}} & \multicolumn{1}{l|}{\cellcolor[HTML]{9AFF99}\begin{tabular}[c]{@{}l@{}}Yes, circular manhole\\ next to tracks.\end{tabular}} & \multicolumn{1}{l|}{\cellcolor[HTML]{9AFF99}\begin{tabular}[c]{@{}l@{}}Manhole cover visible\\ next to tracks.\end{tabular}} & \cellcolor[HTML]{9AFF99}\begin{tabular}[c]{@{}l@{}}Yes, manhole cover next\\ to tracks.\end{tabular} &  \\  
\multicolumn{1}{|c|}{\cellcolor[HTML]{EFEFEF}} & \begin{tabular}[c]{@{}l@{}}Do any vehicles signal before\\ entering the bike lane?\end{tabular} & \multicolumn{1}{l|}{\cellcolor[HTML]{9AFF99}No vehicles signal.} & \multicolumn{1}{l|}{\cellcolor[HTML]{9AFF99}No indication of signaling.} & \multicolumn{1}{l|}{\cellcolor[HTML]{9AFF99}No.} & \multicolumn{1}{l|}{\cellcolor[HTML]{FFCCC9}Vehicles seen using signals.} & \cellcolor[HTML]{FFCCC9}\begin{tabular}[c]{@{}l@{}}Yes, white and black\\ cars signal.\end{tabular} &  \\ \cline{2-7}
\multicolumn{1}{|c|}{\cellcolor[HTML]{EFEFEF}} & \begin{tabular}[c]{@{}l@{}}Does the bike lane have a\\ shared usage sign?\end{tabular} & \multicolumn{1}{l|}{\cellcolor[HTML]{9AFF99}No sign for shared usage.} & \multicolumn{1}{l|}{\cellcolor[HTML]{FFCCC9}\begin{tabular}[c]{@{}l@{}}Clear sign indicating\\ shared use.\end{tabular}} & \multicolumn{1}{l|}{\cellcolor[HTML]{9AFF99}\begin{tabular}[c]{@{}l@{}}No clear signage; bike\\ symbol suggests exclusivity.\end{tabular}} & \multicolumn{1}{l|}{\cellcolor[HTML]{FFCCC9}\begin{tabular}[c]{@{}l@{}}Small, unclear sign\\ present.\end{tabular}} & \cellcolor[HTML]{FFCCC9}Clear sign for shared usage. &  \\ \cline{2-7}
\multicolumn{1}{|c|}{\cellcolor[HTML]{EFEFEF}} & \begin{tabular}[c]{@{}l@{}}Do tram tracks pass through\\ the middle of the road?\end{tabular} & \multicolumn{1}{l|}{\cellcolor[HTML]{9AFF99}\begin{tabular}[c]{@{}l@{}}Yes, tram tracks visible\\ in the road center.\end{tabular}} & \multicolumn{1}{l|}{\cellcolor[HTML]{9AFF99}Tram track in road center.} & \multicolumn{1}{l|}{\cellcolor[HTML]{9AFF99}\begin{tabular}[c]{@{}l@{}}Tram or light rail tracks\\ visible.\end{tabular}} & \multicolumn{1}{l|}{\cellcolor[HTML]{9AFF99}\begin{tabular}[c]{@{}l@{}}Tram tracks in the middle\\ of the road.\end{tabular}} & \cellcolor[HTML]{9AFF99}\begin{tabular}[c]{@{}l@{}}Tram tracks pass\\ through road middle.\end{tabular} &  \\ \cline{2-7}
\multicolumn{1}{|c|}{\multirow{-10}{*}{\rotatebox[origin=c]{90}{\centering {\cellcolor[HTML]{EFEFEF}\textbf{Moderate}}}}}& \begin{tabular}[c]{@{}l@{}}Is there no manhole cover\\ next to the tracks?\end{tabular} & \multicolumn{1}{l|}{\cellcolor[HTML]{FFCCC9}\begin{tabular}[c]{@{}l@{}}Yes, manhole cover next\\ to tracks.\end{tabular}} & \multicolumn{1}{l|}{\cellcolor[HTML]{9AFF99}\begin{tabular}[c]{@{}l@{}}No manhole cover next\\ to tracks.\end{tabular}} & \multicolumn{1}{l|}{\cellcolor[HTML]{9AFF99}\begin{tabular}[c]{@{}l@{}}No manhole cover next\\ to tracks.\end{tabular}} & \multicolumn{1}{l|}{\cellcolor[HTML]{9AFF99}\begin{tabular}[c]{@{}l@{}}No manhole cover next\\ to tracks.\end{tabular}} & \cellcolor[HTML]{FFCCC9}\begin{tabular}[c]{@{}l@{}}Yes, manhole cover next\\ to tracks.\end{tabular} &  \\  \cmidrule{1-7} 
   \cline{1-7}
\multicolumn{1}{|c|}{\cellcolor[HTML]{EFEFEF}} & \begin{tabular}[c]{@{}l@{}}Explain the make of each\\ car turning onto the bike\\ lane.\end{tabular} & \multicolumn{1}{l|}{\cellcolor[HTML]{FFCCC9}\begin{tabular}[c]{@{}l@{}}White sedan, grey SUV,\\ black sedan.\end{tabular}} & \multicolumn{1}{l|}{\cellcolor[HTML]{FFCCC9}\begin{tabular}[c]{@{}l@{}}Various SUV\\ crossovers, sedans.\end{tabular}} & \multicolumn{1}{l|}{\cellcolor[HTML]{FFCCC9}\begin{tabular}[c]{@{}l@{}}White (Mazda/Subaru),\\ black (Toyota/Honda),\\ white sedan.\end{tabular}} & \multicolumn{1}{l|}{\cellcolor[HTML]{FFCCC9}\begin{tabular}[c]{@{}l@{}}White SUV, black\\ sedan, red sports car.\end{tabular}} & \cellcolor[HTML]{9AFF99}\begin{tabular}[c]{@{}l@{}}Sky blue hatchback, maroon\\ sedan, blue sedan,\\ white SUV.\end{tabular} &  \\ \cline{2-7}
\multicolumn{1}{|c|}{\cellcolor[HTML]{EFEFEF}} & \begin{tabular}[c]{@{}l@{}}Which color car is most\\ commonly seen turning into\\ the bike lane?\end{tabular} & \multicolumn{1}{l|}{\cellcolor[HTML]{FFCCC9}White cars.} & \multicolumn{1}{l|}{\cellcolor[HTML]{FFCCC9}\begin{tabular}[c]{@{}l@{}}White cars are the most\\ common.\end{tabular}} & \multicolumn{1}{l|}{\cellcolor[HTML]{FFCCC9}White cars.} & \multicolumn{1}{l|}{\cellcolor[HTML]{FFCCC9}\begin{tabular}[c]{@{}l@{}}White SUV turns into\\ bike lane.\end{tabular}} & \cellcolor[HTML]{FFCCC9}\begin{tabular}[c]{@{}l@{}}Black is most frequent. Blue cars (sky blue, dark\\ blue\end{tabular} &  \\ \cline{2-7}
\multicolumn{1}{|c|}{\cellcolor[HTML]{EFEFEF}} & \begin{tabular}[c]{@{}l@{}}What are the different types\\ of vehicles?\end{tabular} & \multicolumn{1}{l|}{\cellcolor[HTML]{9AFF99}\begin{tabular}[c]{@{}l@{}}Sedans, SUVs, trucks,\\ buses, motorcycles.\end{tabular}} & \multicolumn{1}{l|}{\cellcolor[HTML]{9AFF99}\begin{tabular}[c]{@{}l@{}}Cars, bicycles, motorcycles,\\ pedestrians, bus,\\ truck.\end{tabular}} & \multicolumn{1}{l|}{\cellcolor[HTML]{9AFF99}Sedans, SUVs, bicycles.} & \multicolumn{1}{l|}{\cellcolor[HTML]{9AFF99}Cars, vans, motorcycles.} & \cellcolor[HTML]{9AFF99}\begin{tabular}[c]{@{}l@{}}Cars, trucks, motorcycles,\\ bicycles.\end{tabular} &  \\ \cline{2-7}
\multicolumn{1}{|c|}{\cellcolor[HTML]{EFEFEF}} & \begin{tabular}[c]{@{}l@{}}What is the intersection design\\ causing cars to turn into\\ bike lanes?\end{tabular} & \multicolumn{1}{l|}{\cellcolor[HTML]{9AFF99}\begin{tabular}[c]{@{}l@{}}Shared lane, poorly\\ marked.\end{tabular}} & \multicolumn{1}{l|}{\cellcolor[HTML]{9AFF99}\begin{tabular}[c]{@{}l@{}}Bike-friendly, shared\\ for cars and bikes.\end{tabular}} & \multicolumn{1}{l|}{\cellcolor[HTML]{9AFF99}\begin{tabular}[c]{@{}l@{}}Poorly marked lanes\\ causing confusion.\end{tabular}} & \multicolumn{1}{l|}{\cellcolor[HTML]{FFCCC9}No specific design mentioned.} & \cellcolor[HTML]{9AFF99}\begin{tabular}[c]{@{}l@{}}Shared road space due\\ to narrow lane.\end{tabular} &  \\ \cline{2-7}
\multicolumn{1}{|c|}{\cellcolor[HTML]{EFEFEF}} & \begin{tabular}[c]{@{}l@{}}Do pedestrians at the start\\ return at the end?\end{tabular} & \multicolumn{1}{l|}{\cellcolor[HTML]{9AFF99}\begin{tabular}[c]{@{}l@{}}No pedestrians at start\\ or end.\end{tabular}} & \multicolumn{1}{l|}{\cellcolor[HTML]{FFCCC9}\begin{tabular}[c]{@{}l@{}}Pedestrians at start, returning\\ at end.\end{tabular}} & \multicolumn{1}{l|}{\cellcolor[HTML]{9AFF99}\begin{tabular}[c]{@{}l@{}}No, different pedestrians\\ at end.\end{tabular}} & \multicolumn{1}{l|}{\cellcolor[HTML]{9AFF99}\begin{tabular}[c]{@{}l@{}}No, initial pedestrians\\ do not return.\end{tabular}} & \cellcolor[HTML]{9AFF99}\begin{tabular}[c]{@{}l@{}}No return of initial\\ pedestrians.\end{tabular} &  \\ \cline{2-7}
\multicolumn{1}{|c|}{\multirow{-10}{*}{\rotatebox[origin=c]{90}{\centering {\cellcolor[HTML]{EFEFEF}\textbf{Complex}}}}} & \begin{tabular}[c]{@{}l@{}}Which color car is not most\\ commonly seen turning into\\ the bike lane?\end{tabular} & \multicolumn{1}{l|}{\cellcolor[HTML]{FFCCC9}\begin{tabular}[c]{@{}l@{}}Any color other than\\ red.\end{tabular}} & \multicolumn{1}{l|}{\cellcolor[HTML]{FFCCC9}\begin{tabular}[c]{@{}l@{}}Car not turning into bike\\ lane.\end{tabular}} & \multicolumn{1}{l|}{\cellcolor[HTML]{9AFF99}Blue car is least common.} & \multicolumn{1}{l|}{\cellcolor[HTML]{FFCCC9}\begin{tabular}[c]{@{}l@{}}Color not shown in\\ video.\end{tabular}} & \cellcolor[HTML]{9AFF99}Black car is least frequent. 
& \\
\cmidrule{1-7} 
   \cline{1-7}
\end{tabular}
}\end{table}

\begin{table}[h]
\caption{Question sets used on synthetic traffic video sequence 3}
\label{allQuestionsSeq3}
\resizebox{\textwidth}{!}{%
\begin{tabular}{
>{\columncolor[HTML]{FFFFFF}}c |
>{\columncolor[HTML]{FFFFFF}}l l
>{\columncolor[HTML]{FFCCC9}}l 
>{\columncolor[HTML]{FFCCC9}}l 
>{\columncolor[HTML]{FFCCC9}}l l|}
\cline{2-7}
\multicolumn{1}{l|}{\cellcolor[HTML]{FFFFFF}} &
  \multicolumn{6}{c|}{\cellcolor[HTML]{FFFFFF}\includegraphics[width=1.5\textwidth]{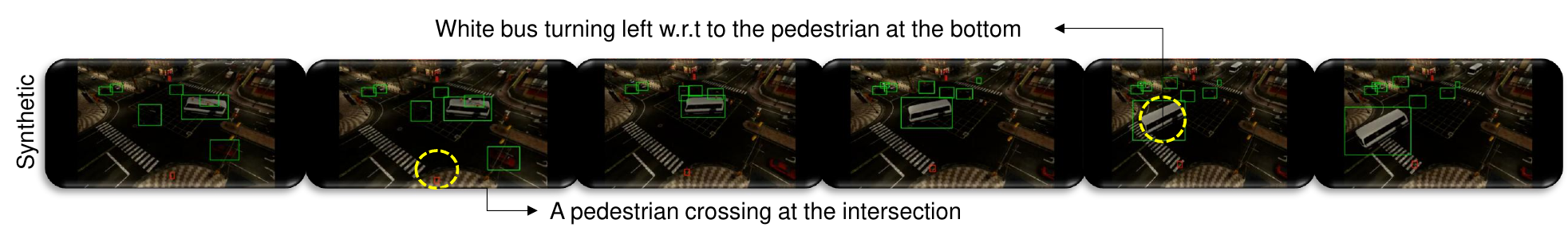}}  \\  \cline{2-7} 
\multicolumn{1}{l|}{\cellcolor[HTML]{FFFFFF}} &
  \multicolumn{1}{c|}{\cellcolor[HTML]{EFEFEF}} &
  \multicolumn{5}{c|}{\cellcolor[HTML]{EFEFEF}\textbf{Video QA model Answers}} \\ \cline{3-7} 
\multicolumn{1}{l|}{\multirow{-3}{*}{\cellcolor[HTML]{FFFFFF}}} &
  \multicolumn{1}{c|}{\multirow{-2}{*}{\cellcolor[HTML]{EFEFEF}\textbf{Questions}}} &
  \multicolumn{1}{c|}{\cellcolor[HTML]{EFEFEF}\textbf{ChatGPT}} &
  \multicolumn{1}{c|}{\cellcolor[HTML]{EFEFEF}\textbf{LLaVa}} &
  \multicolumn{1}{c|}{\cellcolor[HTML]{EFEFEF}\textbf{Gemini}} &
  \multicolumn{1}{c|}{\cellcolor[HTML]{EFEFEF}\textbf{InternVL}} &
  \multicolumn{1}{c|}{\cellcolor[HTML]{EFEFEF}\textbf{VideoLLaMA-2}} \\ \hline \hline
\multicolumn{1}{|c|}{\cellcolor[HTML]{EFEFEF}} &
  \multicolumn{1}{l|}{\cellcolor[HTML]{FFFFFF}\begin{tabular}[c]{@{}l@{}}Is the pedestrian using a \\ crosswalk?\end{tabular}} &
  \multicolumn{1}{l|}{\cellcolor[HTML]{9AFF99}\begin{tabular}[c]{@{}l@{}}Yes, the pedestrian is using \\ a crosswalk.\end{tabular}} &
  \multicolumn{1}{l|}{\cellcolor[HTML]{9AFF99}\begin{tabular}[c]{@{}l@{}}The pedestrian is using a crosswalk, \\ marked by white lines.\end{tabular}} &
  \multicolumn{1}{l|}{\cellcolor[HTML]{9AFF99}\begin{tabular}[c]{@{}l@{}}Yes, the pedestrian uses \\ the crosswalk.\end{tabular}} &
  \multicolumn{1}{l|}{\cellcolor[HTML]{9AFF99}\begin{tabular}[c]{@{}l@{}}Yes, the pedestrian is using \\ a crosswalk to cross.\end{tabular}} &
  \cellcolor[HTML]{9AFF99}\begin{tabular}[c]{@{}l@{}}Yes, the pedestrian is using a \\ crosswalk.\end{tabular} \\ \cline{2-7} 
\multicolumn{1}{|c|}{\cellcolor[HTML]{EFEFEF}} &
  \multicolumn{1}{l|}{\cellcolor[HTML]{FFFFFF}\begin{tabular}[c]{@{}l@{}}How many cars can you see at \\ the start?\end{tabular}} &
  \multicolumn{1}{l|}{\cellcolor[HTML]{FFCCC9}\begin{tabular}[c]{@{}l@{}}Three cars are visible at \\ the start.\end{tabular}} &
  \multicolumn{1}{l|}{\cellcolor[HTML]{FFCCC9}Two cars are visible} &
  \multicolumn{1}{l|}{\cellcolor[HTML]{FFCCC9}\begin{tabular}[c]{@{}l@{}}Five cars, including a bus, \\ are visible initially.\end{tabular}} &
  \multicolumn{1}{l|}{\cellcolor[HTML]{FFCCC9}Three cars are visible.} &
  \cellcolor[HTML]{FFCCC9}Two cars are visible. \\ \cline{2-7} 
\multicolumn{1}{|c|}{\cellcolor[HTML]{EFEFEF}} &
  \multicolumn{1}{l|}{\cellcolor[HTML]{FFFFFF}Is it day or night?} &
  \multicolumn{1}{l|}{\cellcolor[HTML]{FFCCC9}It is daytime.} &
  \multicolumn{1}{l|}{\cellcolor[HTML]{9AFF99}It is nighttime in the image.} &
  \multicolumn{1}{l|}{\cellcolor[HTML]{9AFF99}It is nighttime.} &
  \multicolumn{1}{l|}{\cellcolor[HTML]{FFCCC9}\begin{tabular}[c]{@{}l@{}}It appears to be twilight, not \\ full daylight.\end{tabular}} &
  \cellcolor[HTML]{9AFF99}It is night. \\ \cline{2-7} 
\multicolumn{1}{|c|}{\cellcolor[HTML]{EFEFEF}} &
  \multicolumn{1}{l|}{\cellcolor[HTML]{FFFFFF}\begin{tabular}[c]{@{}l@{}}How many yellow cars \\ are there?\end{tabular}} &
  \multicolumn{1}{l|}{\cellcolor[HTML]{9AFF99}There is one yellow car.} &
  \multicolumn{1}{l|}{\cellcolor[HTML]{9AFF99}There are two yellow cars} &
  \multicolumn{1}{l|}{\cellcolor[HTML]{9AFF99}One yellow car (a taxi) is visible.} &
  \multicolumn{1}{l|}{\cellcolor[HTML]{9AFF99}There is one yellow car.} &
  \cellcolor[HTML]{9AFF99}\begin{tabular}[c]{@{}l@{}}There are a couple of yellow \\ cars.\end{tabular} \\ \cline{2-7} 
\multicolumn{1}{|c|}{\cellcolor[HTML]{EFEFEF}} &
  \multicolumn{1}{l|}{\cellcolor[HTML]{FFFFFF}\begin{tabular}[c]{@{}l@{}}How many traffic lights \\ are visible?\end{tabular}} &
  \multicolumn{1}{l|}{\cellcolor[HTML]{FFCCC9}Two traffic lights are visible.} &
  \multicolumn{1}{l|}{\cellcolor[HTML]{FFCCC9}Four traffic lights are visible} &
  \multicolumn{1}{l|}{\cellcolor[HTML]{FFCCC9}Two traffic lights are partially visible.} &
  \multicolumn{1}{l|}{\cellcolor[HTML]{FFCCC9}One traffic light is visible} &
  \cellcolor[HTML]{FFCCC9}\begin{tabular}[c]{@{}l@{}}Multiple traffic lights are \\ visible but unclear count.\end{tabular} \\ \cline{2-7} 
\multicolumn{1}{|c|}{\multirow{-10}{*} {\rotatebox[origin=c]{90}{\centering {\cellcolor[HTML]{EFEFEF}\textbf{Easy}}}}} &
  \multicolumn{1}{l|}{\cellcolor[HTML]{FFFFFF}\begin{tabular}[c]{@{}l@{}}Is the pedestrian not \\ using a crosswalk?\end{tabular}} &
  \multicolumn{1}{l|}{\cellcolor[HTML]{FFCCC9}\begin{tabular}[c]{@{}l@{}}The pedestrian was using \\ a crosswalk.\end{tabular}} &
  \multicolumn{1}{l|}{\cellcolor[HTML]{FFCCC9}\begin{tabular}[c]{@{}l@{}}The pedestrian is using the crosswalk \\ as intended.\end{tabular}} &
  \multicolumn{1}{l|}{\cellcolor[HTML]{FFCCC9}The pedestrian is using a crosswalk} &
  \multicolumn{1}{l|}{\cellcolor[HTML]{FFCCC9}\begin{tabular}[c]{@{}l@{}}Detailed analysis indicates \\ pedestrian use of crosswalk.\end{tabular}} &
  \cellcolor[HTML]{9AFF99}\begin{tabular}[c]{@{}l@{}}Yes, the pedestrian is not \\ using a crosswalk.\end{tabular} \\ \hline \hline
\multicolumn{1}{|c|}{\cellcolor[HTML]{EFEFEF}} &
  \multicolumn{1}{l|}{\cellcolor[HTML]{FFFFFF}\begin{tabular}[c]{@{}l@{}}How many pedestrians \\ cross the road in total?\end{tabular}} &
  \multicolumn{1}{l|}{\cellcolor[HTML]{FFCCC9}\begin{tabular}[c]{@{}l@{}}One pedestrian crosses \\ the road.\end{tabular}} &
  \multicolumn{1}{l|}{\cellcolor[HTML]{FFCCC9}Ten pedestrians are visible.} &
  \multicolumn{1}{l|}{\cellcolor[HTML]{FFCCC9}One pedestrian crosses the road.} &
  \multicolumn{1}{l|}{\cellcolor[HTML]{FFCCC9}\begin{tabular}[c]{@{}l@{}}Four pedestrians cross the \\ road in total\end{tabular}} &
  \cellcolor[HTML]{FFCCC9}\begin{tabular}[c]{@{}l@{}}Four pedestrians are seen \\ crossing.\end{tabular} \\ \cline{2-7} 
\multicolumn{1}{|c|}{\cellcolor[HTML]{EFEFEF}} &
  \multicolumn{1}{l|}{\cellcolor[HTML]{FFFFFF}\begin{tabular}[c]{@{}l@{}}How many vehicles \\ stopped for pedestrians?\end{tabular}} &
  \multicolumn{1}{l|}{\cellcolor[HTML]{9AFF99}\begin{tabular}[c]{@{}l@{}}Two vehicles stopped for \\ the pedestrian.\end{tabular}} &
  \multicolumn{1}{l|}{\cellcolor[HTML]{FFCCC9}Four vehicles stopped.} &
  \multicolumn{1}{l|}{\cellcolor[HTML]{FFCCC9}One vehicle (the white bus) stops.} &
  \multicolumn{1}{l|}{\cellcolor[HTML]{9AFF99}Two vehicles stopped.} &
  \cellcolor[HTML]{FFCCC9}\begin{tabular}[c]{@{}l@{}}Three vehicles stopped for \\ pedestrians.\end{tabular} \\ \cline{2-7} 
\multicolumn{1}{|c|}{\cellcolor[HTML]{EFEFEF}} &
  \multicolumn{1}{l|}{\cellcolor[HTML]{FFFFFF}\begin{tabular}[c]{@{}l@{}}Were the pedestrians crossing \\ from left to right or right to left?\end{tabular}} &
  \multicolumn{1}{l|}{\cellcolor[HTML]{9AFF99}\begin{tabular}[c]{@{}l@{}}The pedestrian was crossing \\ from left to right.\end{tabular}} &
  \multicolumn{1}{l|}{\cellcolor[HTML]{9AFF99}\begin{tabular}[c]{@{}l@{}}The pedestrian is crossing from\\  left to right.\end{tabular}} &
  \multicolumn{1}{l|}{\cellcolor[HTML]{FFCCC9}\begin{tabular}[c]{@{}l@{}}The pedestrian crosses from bottom \\ left to top right.\end{tabular}} &
  \multicolumn{1}{l|}{\cellcolor[HTML]{FFCCC9}\begin{tabular}[c]{@{}l@{}}The pedestrian was crossing \\ from right to left.\end{tabular}} &
  \cellcolor[HTML]{FFCCC9}\begin{tabular}[c]{@{}l@{}}The pedestrian was crossing \\ from right to left.\end{tabular} \\ \cline{2-7} 
\multicolumn{1}{|c|}{\cellcolor[HTML]{EFEFEF}} &
  \multicolumn{1}{l|}{\cellcolor[HTML]{FFFFFF}\begin{tabular}[c]{@{}l@{}}Which car arrived first at the \\ intersection?\end{tabular}} &
  \multicolumn{1}{l|}{\cellcolor[HTML]{FFCCC9}A sedan arrived first.} &
  \multicolumn{1}{l|}{\cellcolor[HTML]{FFCCC9}The car on the right arrived first.} &
  \multicolumn{1}{l|}{\cellcolor[HTML]{9AFF99}The white bus arrives first.} &
  \multicolumn{1}{l|}{\cellcolor[HTML]{FFCCC9}\begin{tabular}[c]{@{}l@{}}Red car arrives first after \\ detailed breakdown.\end{tabular}} &
  \cellcolor[HTML]{9AFF99}The white car arrived first. \\ \cline{2-7} 
\multicolumn{1}{|c|}{\cellcolor[HTML]{EFEFEF}} &
  \multicolumn{1}{l|}{\cellcolor[HTML]{FFFFFF}\begin{tabular}[c]{@{}l@{}}Was the pedestrian in the middle \\ of the road at any point?\end{tabular}} &
  \multicolumn{1}{l|}{\cellcolor[HTML]{9AFF99}\begin{tabular}[c]{@{}l@{}}Yes, the pedestrian was in \\ the middle briefly.\end{tabular}} &
  \multicolumn{1}{l|}{\cellcolor[HTML]{FFCCC9}\begin{tabular}[c]{@{}l@{}}No, the pedestrian was not in the \\ middle.\end{tabular}} &
  \multicolumn{1}{l|}{\cellcolor[HTML]{9AFF99}\begin{tabular}[c]{@{}l@{}}Yes, the pedestrian is in the middle \\ while crossing.\end{tabular}} &
  \multicolumn{1}{l|}{\cellcolor[HTML]{9AFF99}\begin{tabular}[c]{@{}l@{}}The pedestrian was briefly \\ in the middle.\end{tabular}} &
  \cellcolor[HTML]{9AFF99}\begin{tabular}[c]{@{}l@{}}Yes, the pedestrian was in \\ the middle.\end{tabular} \\ \cline{2-7} 
\multicolumn{1}{|c|}{\multirow{-10}{*} {\rotatebox[origin=c]{90}{\centering {\cellcolor[HTML]{EFEFEF}\textbf{Moderate}}}}} &
  \multicolumn{1}{l|}{\cellcolor[HTML]{FFFFFF}\begin{tabular}[c]{@{}l@{}}How many pedestrians did not \\ cross the road?\end{tabular}} &
  \multicolumn{1}{l|}{\cellcolor[HTML]{FFCCC9}\begin{tabular}[c]{@{}l@{}}Only 1 pedestrian crossed, \\ so 0 did not cross.\end{tabular}} &
  \multicolumn{1}{l|}{\cellcolor[HTML]{FFCCC9}Ten pedestrians did not cross.} &
  \multicolumn{1}{l|}{\cellcolor[HTML]{9AFF99}No pedestrians did not cross.} &
  \multicolumn{1}{l|}{\cellcolor[HTML]{FFCCC9}\begin{tabular}[c]{@{}l@{}}Two pedestrians did \\ not cross.\end{tabular}} &
  \cellcolor[HTML]{FFCCC9}Four pedestrians did not cross. \\ \hline \hline 
\multicolumn{1}{|c|}{\cellcolor[HTML]{EFEFEF}} &
  \multicolumn{1}{l|}{\cellcolor[HTML]{FFFFFF}\begin{tabular}[c]{@{}l@{}}Impact of pedestrian position \\ on vehicle flow?\end{tabular}} &
  \multicolumn{1}{l|}{\cellcolor[HTML]{9AFF99}\begin{tabular}[c]{@{}l@{}}The pedestrian slowed \\ down vehicle flow.\end{tabular}} &
  \multicolumn{1}{l|}{\cellcolor[HTML]{FFCCC9}\begin{tabular}[c]{@{}l@{}}Pedestrian position does not impact \\ vehicle flow.\end{tabular}} &
  \multicolumn{1}{l|}{\cellcolor[HTML]{9AFF99}Pedestrian causes white bus to stop.} &
  \multicolumn{1}{l|}{\cellcolor[HTML]{9AFF99}\begin{tabular}[c]{@{}l@{}}Pedestrian presence causes \\ potential congestion.\end{tabular}} &
  \cellcolor[HTML]{FFCCC9}No significant impact observed. \\ \cline{2-7} 
\multicolumn{1}{|c|}{\cellcolor[HTML]{EFEFEF}} &
  \multicolumn{1}{l|}{\cellcolor[HTML]{FFFFFF}\begin{tabular}[c]{@{}l@{}}How many vehicles stopped \\ after turning?\end{tabular}} &
  \multicolumn{1}{l|}{\cellcolor[HTML]{9AFF99}\begin{tabular}[c]{@{}l@{}}One vehicle stopped\\ after turning.\end{tabular}} &
  \multicolumn{1}{l|}{\cellcolor[HTML]{FFCCC9}Unable to determine exact count.} &
  \multicolumn{1}{l|}{\cellcolor[HTML]{FFCCC9}Zero vehicles stopped after turning.} &
  \multicolumn{1}{l|}{\cellcolor[HTML]{FFCCC9}\begin{tabular}[c]{@{}l@{}}Two vehicles stopped \\ after turning\end{tabular}} &
  \cellcolor[HTML]{9AFF99}\begin{tabular}[c]{@{}l@{}}One vehicle stopped after \\ turning.\end{tabular} \\ \cline{2-7} 
\multicolumn{1}{|c|}{\cellcolor[HTML]{EFEFEF}} &
  \multicolumn{1}{l|}{\cellcolor[HTML]{FFFFFF}\begin{tabular}[c]{@{}l@{}}Vehicles in the intersection \\ after pedestrians crossed?\end{tabular}} &
  \multicolumn{1}{l|}{\cellcolor[HTML]{9AFF99}\begin{tabular}[c]{@{}l@{}}Two vehicles were in \\ the middle\end{tabular}} &
  \multicolumn{1}{l|}{\cellcolor[HTML]{FFCCC9}Not clear from the image.} &
  \multicolumn{1}{l|}{\cellcolor[HTML]{FFCCC9}One yellow taxi.} &
  \multicolumn{1}{l|}{\cellcolor[HTML]{FFCCC9}\begin{tabular}[c]{@{}l@{}}Three vehicles in the \\ middle.\end{tabular}} &
  \cellcolor[HTML]{9AFF99}\begin{tabular}[c]{@{}l@{}}Two vehicles were in the \\ middle.\end{tabular} \\ \cline{2-7} 
\multicolumn{1}{|c|}{\cellcolor[HTML]{EFEFEF}} &
  \multicolumn{1}{l|}{\cellcolor[HTML]{FFFFFF}\begin{tabular}[c]{@{}l@{}}Longest gap between vehicle \\ movement and pedestrian crossing?\end{tabular}} &
  \multicolumn{1}{l|}{\cellcolor[HTML]{FFCCC9}Approximately 3 seconds.} &
  \multicolumn{1}{l|}{\cellcolor[HTML]{FFCCC9}Longest gap is 10 seconds.} &
  \multicolumn{1}{l|}{\cellcolor[HTML]{FFCCC9}\begin{tabular}[c]{@{}l@{}}Longest gap after bus movement, \\ no vehicles for pedestrians\end{tabular}} &
  \multicolumn{1}{l|}{\cellcolor[HTML]{FFCCC9}\begin{tabular}[c]{@{}l@{}}Detailed frame-by-frame \\ analysis shows multiple gaps.\end{tabular}} &
  \cellcolor[HTML]{9AFF99}Longest gap of 2 to 3 seconds. \\ \cline{2-7} 
\multicolumn{1}{|c|}{\cellcolor[HTML]{EFEFEF}} &
  \multicolumn{1}{l|}{\cellcolor[HTML]{FFFFFF}Was traffic light timing optimal?} &
  \multicolumn{1}{l|}{\cellcolor[HTML]{FFCCC9}\begin{tabular}[c]{@{}l@{}}Traffic light timing was \\ moderately effective.\end{tabular}} &
  \multicolumn{1}{l|}{\cellcolor[HTML]{FFCCC9}Analysis not possible from image.} &
  \multicolumn{1}{l|}{\cellcolor[HTML]{FFCCC9}\begin{tabular}[c]{@{}l@{}}Traffic lights function as expected, \\ but full view not seen.\end{tabular}} &
  \multicolumn{1}{l|}{\cellcolor[HTML]{9AFF99}\begin{tabular}[c]{@{}l@{}}Traffic light timing could \\ be more optimal.\end{tabular}} &
  \cellcolor[HTML]{9AFF99}\begin{tabular}[c]{@{}l@{}}Traffic light timing appeared \\ optimal.\end{tabular} \\ \cline{2-7} 
\multicolumn{1}{|c|}{\multirow{-12}{*} {\rotatebox[origin=c]{90}{\centering {\cellcolor[HTML]{EFEFEF}\textbf{Complex}}}}} &
  \multicolumn{1}{l|}{\cellcolor[HTML]{FFFFFF}\begin{tabular}[c]{@{}l@{}}Vehicles not in the middle of the \\ intersection  after pedestrians crossed?\end{tabular}} &
  \multicolumn{1}{l|}{\cellcolor[HTML]{FFCCC9}\begin{tabular}[c]{@{}l@{}}All vehicles not in the \\ middle post-crossing.\end{tabular}} &
  \multicolumn{1}{l|}{\cellcolor[HTML]{FFCCC9}Pedestrian uses crosswalk as intended.} &
  \multicolumn{1}{l|}{\cellcolor[HTML]{FFCCC9}Five vehicles not in the middle.} &
  \multicolumn{1}{l|}{\cellcolor[HTML]{FFCCC9}Two vehicles not in the middle.} &
  \cellcolor[HTML]{FFCCC9}Two vehicles not in the middle. \\ \hline \hline
\end{tabular}%
}
\end{table}

\end{document}